\documentclass[runningheads]{llncs}
\usepackage[T1]{fontenc}
\usepackage{graphicx}
\usepackage{booktabs}
\usepackage[misc]{ifsym}
\newcommand{\corr}{(\Letter)}
\usepackage{mwe}

\usepackage[utf8]{inputenc} 
\usepackage[T1]{fontenc}    
\usepackage{hyperref}       
\usepackage{breakurl}
\usepackage{booktabs}       
\usepackage{amsfonts}       
\usepackage{nicefrac}       
\usepackage{microtype}      
\usepackage{xcolor}         
\usepackage[noend]{algorithm2e}
\RestyleAlgo{ruled}
\SetKwComment{Comment}{/}{}
\usepackage{array}
\usepackage{multirow}

\usepackage{stmaryrd}
\usepackage{amsmath}
\usepackage{dsfont}
\usepackage{graphicx}
\usepackage{tikz,graphics,color,float,epsf,caption,subcaption}
\usepackage{hhline}

\begin{document}

\title{Hyperbolic Delaunay Geometric Alignment}


\author{Aniss Aiman Medbouhi\corr \orcidID{0000-0002-6649-3325} \and
Giovanni Luca Marchetti\orcidID{0009-0004-8248-229X} \and
Vladislav Polianskii\orcidID{0000-0001-9805-0388} \and Alexander Kravberg\orcidID{0000-0002-9001-7708} \and Petra Poklukar\orcidID{0000-0001-6920-5109} \and Anastasia Varava\orcidID{0000-0002-0900-1523} \and Danica Kragic \orcidID{0000-0003-2965-2953}}

\authorrunning{A.A. Medbouhi et al.}

\institute{Division of Robotics Perception and Learning, Department of Intelligent Systems, School of Electrical Engineering and Computer Science,\\
KTH Royal Institute of Technology, 100 44 Stockholm, Sweden\\
\email{medbouhi@kth.se}
}

\maketitle   

\begin{abstract}
Hyperbolic machine learning is an emerging field aimed at representing data with a hierarchical structure. However, there is a lack of tools for evaluation and analysis of the resulting hyperbolic data representations. To this end, we propose Hyperbolic Delaunay Geometric Alignment (HyperDGA) -- a similarity score for comparing datasets in a hyperbolic space. The core idea is counting the edges of the hyperbolic Delaunay graph connecting datapoints across the given sets. We provide an empirical investigation on synthetic and real-life biological data and demonstrate that HyperDGA outperforms the hyperbolic version of classical distances between sets. Furthermore, we showcase the potential of HyperDGA for evaluating latent representations inferred by a Hyperbolic Variational Auto-Encoder. 

\end{abstract}

\section{Introduction}

\emph{Hyperbolic} geometry is a non-Euclidean geometry characterized by constant negative curvature \cite{Beltrami1868_EssayInterpretationNonEuclideanGeometry}, which is particularly suitable for low-dimensional tree embedding. In contrast to the Euclidean case, a hyperbolic space requires only two dimensions to embed any tree with arbitrary low distortion due to the exponential volume growth away from the origin \cite{Sarkar2012_LowDistortionEmbeddingTreesHyperbolicPlane}. 
This has motivated the use of hyperbolic geometry in machine learning for representing data that exhibit a \emph{hierarchical} structure. Examples of such data include geographic
communication networks \cite{GeographicRoutingUsingHyperbolic}, internet networks \cite{InternetHyperbolicMapping}, words in a natural language \cite{Nickel_2017_PoincareEmbedding,Tifrea2019PoincareGlove}, or single-cell data in biology \cite{Nickel_2020_SingleCellPoincareEmbedding}. Recent work in dimensionality reduction \cite{Sala2018_RepresentationTradeoffsHyperbolicEmbeddings,Chami_2021_HoroPCA,Guo_2022_CO-SNE} and generative modeling \cite{Mathieu2019_PoincareVAE,Nagano2019_WrappedNormalDistributionHyperbolicSpaceGradientBasedLearning} has shown that performance in downstream tasks is improved when hierarchical data is represented in a hyperbolic space.

\begin{figure}[t]
    \centering
        \includegraphics[width=1\textwidth]{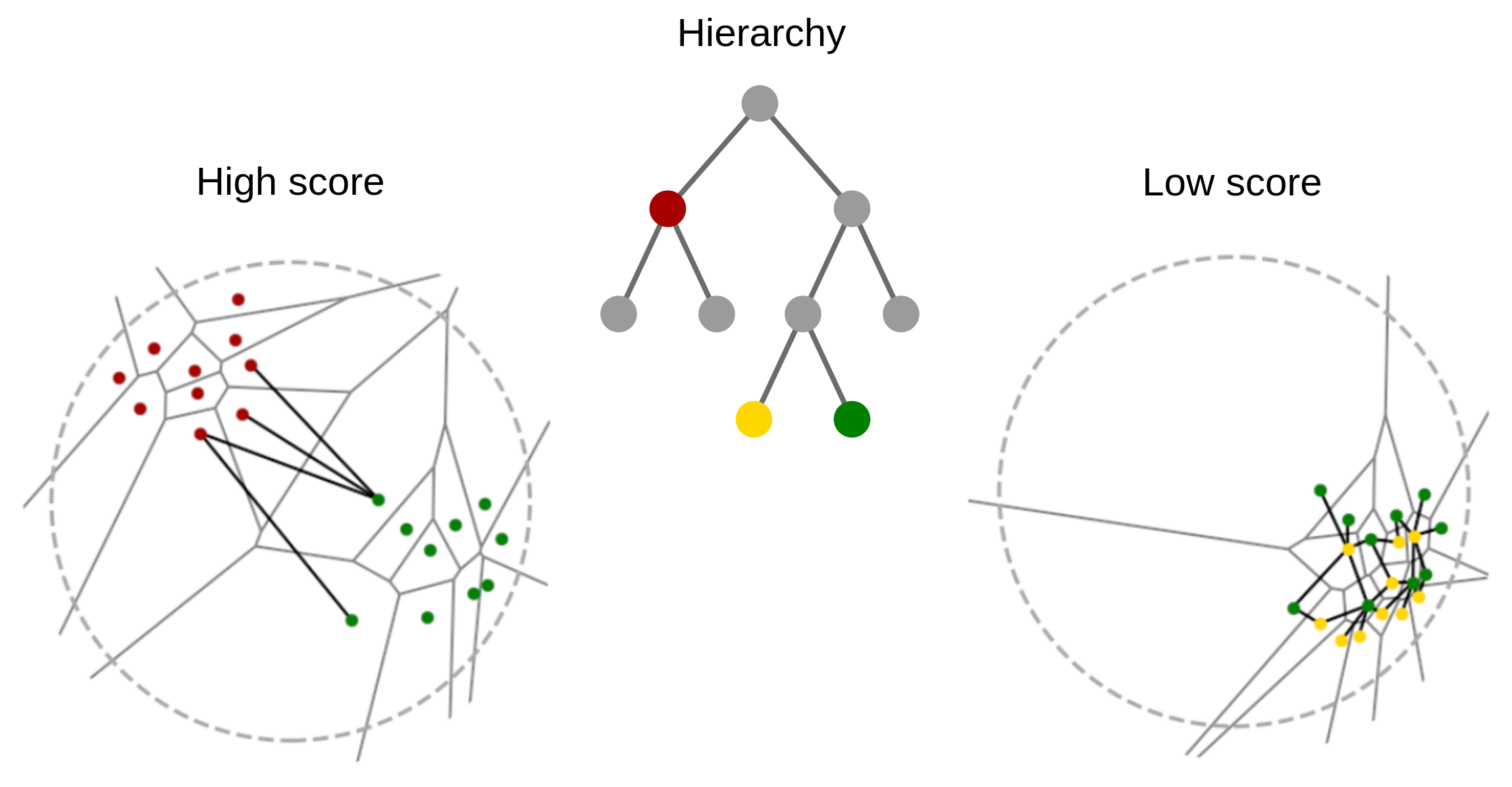}
        \caption{Illustration of HyperDGA for a hyperbolic representation of hierarchical data. Our score counts heterogeneous edges (black) of the Delaunay graph, which is dual to the hyperbolic Voronoi diagram (gray). Data closer in the hierarchy (green \& yellow) have lower scores than data further apart (red \& green).}
        \label{fig:cover}
\end{figure}

Despite the abundance of hyperbolic machine learning models, there is a lack of methods for analyzing hyperbolic embeddings or for evaluating hyperbolic data representations independently of any downstream task. This raises the need to design evaluation metrics suitable for hyperbolic geometry. To this end, we propose Hyperbolic Delaunay Geometric Alignment (HyperDGA) -- a similarity score between two sets embedded in a hyperbolic space. Our method is based on a hyperbolic Delaunay triangulation that takes into account the topology and geometry of the data representation. Inspired by a recently-introduced score deemed Delaunay Component Analysis (DCA) \cite{Poklukar2021_GeomCA,Poklukar2022_DCA}, HyperDGA relies on counting the edges of the Delaunay graph that are \emph{heterogeneous} i.e., connecting points across the given sets -- see Figure \ref{fig:cover} for an illustration.

We validate HyperDGA empirically on synthetic and real-life datasets represented in a hyperbolic space, and compare it to hyperbolic versions of standard distances for sets -- the Chamfer and the Wasserstein distances. Namely, we investigate correlation of the metrics with noise in a synthetic dataset simulating protein evolution, and evaluate the latent representation of a Hyperbolic Variational Auto-Encoder trained on the same dataset. Moreover, we compare the measured distances on real-life biological data representing cells. We demonstrate that HyperDGA outperforms the Chamfer distance, and performs comparably or better than the Wasserstein distance.

In summary, our contributions include a geometric similarity score between two sets in a hyperbolic space, and an experimental investigation showcasing how HyperDGA can be used for hyperbolic data analysis and evaluation of hyperbolic representations. The code for HyperDGA and the experiments is accessible online: \href{https://github.com/anissmedbouhi/HyperDGA}{https://github.com/anissmedbouhi/HyperDGA}.

\section{Related Work}

Since HyperDGA is designed for data analysis of hyperbolic embeddings, we start by reviewing hyperbolic dimensionality reduction techniques, as well as hyperbolic Variational Auto-Encoders (VAEs). Moreover, we review recent methods for evaluating data representations in the Euclidean space, since this has not been investigated yet in the hyperbolic case.

\subsubsection{Hyperbolic Dimensionality Reduction}
Classical dimensionality reduction methods have been extended to the hyperbolic case, for example Multidimensional Scaling \cite{Sala2018_RepresentationTradeoffsHyperbolicEmbeddings}, $t$-Distributed Stochastic Neighbor Embeddings \cite{Guo_2022_CO-SNE}, Principal Component Analysis \cite{Fletcher2004PrincipalGeodesicAnalysis,Chami_2021_HoroPCA}, and contrastive learning \cite{Nickel_2017_PoincareEmbedding,chamberlain2017NeuralEmbeddingHyperbolic}. All these methods embed data in a lower-dimensional hyperbolic space in order to exploit the inherent hierarchical structure. HyperDGA is applicable on top of these dimensionality reduction techniques to analyze the resulting hyperbolic data representations.

\subsubsection{Hyperbolic VAEs}
Hyperbolic VAEs were first proposed by \cite{Mathieu2019_PoincareVAE} and \cite{Nagano2019_WrappedNormalDistributionHyperbolicSpaceGradientBasedLearning} as hyperbolic analogues of the VAE \cite{VAE,VAE_rezende}. These models differ in two aspects: the model of hyperbolic geometry for the latent space, and the type of hyperbolic Gaussian distribution deployed. Specifically, the hyperbolic models involved are the Poincaré ball model and the hyperboloid model respectively, while the distributions are the Riemannian normal and the wrapped normal, respectively. Mixed-Curvature VAEs \cite{Skopek2020_Mixed-curvatureVAE} can be seen as a generalization of the above-cited hyperbolic VAEs. The main difference is that the latent space consists of a product of spaces of constant curvature selected among the Euclidean, spherical, or hyperbolic one. Thus, the latent space has a non-constant curvature, resulting in more flexibility to represent the data.

\subsubsection{Evaluation of Euclidean Data Representations}
Several methods have been proposed to compare data representations in terms of their geometric and topological structure. For example, Geometry Score \cite{Khrulkov2018_GeometryScore} computes a topological approximation of the given datasets. This results in a the score that is robust to noise, but is computationally expensive and requires tuning several hyperparameters. Similarly, Improved Precision and Recall Metric (IPR) \cite{Kynkaanniemi2019_ImprovedPrescisionRecall} approximates the underlying data manifold by a union of hyperspheres, but only contains a single hyperparameter controlling their sizes. IPR consists of two scores \emph{precision} and \emph{recall}, counting the number of points of a dataset in the other dataset's approximated manifold. Geometric Component Analysis (GeomCA) \cite{Poklukar2021_GeomCA} is inspired by IPR and similarly provides local and global geometric scores. GeomCA approximates the manifold via an $\epsilon$-proximity graph connecting neighboring datapoints, where $\epsilon$ is a hyperparameter controlling the neighbor size. The latter is difficult to tune because clusters of different scales may appear in the data. DCA \cite{Poklukar2022_DCA} amends to this shortcoming by deploying the same evaluation scores while approximating the data manifold via the 1-skeleton of the Delaunay triangulation. In summary, the above methods are designed specifically for Euclidean data representations and cannot be applied directly to different geometries. On the contrary, HyperDGA extends DCA to the hyperbolic space, and therefore is suitable for analyzing and evaluating hyperbolic data representations.

\section{Background}
HyperDGA is based on the computation of a hyperbolic Delaunay graph. The latter captures geometric and topological information in data, which motivates our similarity score. However, a core challenge is computing the Delaunay graph in a hyperbolic space. To this end, we deploy the \emph{Klein-Beltrami model} of hyperbolic geometry, where the geodesics are straight lines. As a consequence, the computation of the  Delaunay graph can be reduced to the Euclidean case, which in turn ca be addressed via standard algorithms from computational geometry. In what follows, we introduce the necessary geometric background on Delaunay triangulations and the Klein-Beltrami model. 

\subsection{Voronoi Cells and Delaunay Graph}

A \emph{Delaunay triangulation} is a collection of simplices in a metric space that is \emph{dual} to the \emph{Voronoi diagram} \cite{book_AlgorithmicGeometry}. The latter partitions the space into regions called \emph{Voronoi cells}. Intuitively, given a set of points, each one defines a Voronoi cell with the property that points in $\mathbb{R}^n$ belong to the cell of their nearest neighbor. More formally, let $\mathcal{X}$ be a metric space with distance function $d\colon \mathcal{X} \times \mathcal{X} \to \mathbb{R}_{\geq 0}$ and let $P \subseteq \mathcal{X}$ be a finite subset.

\begin{definition}\label{def:DelaunayGraph}
The \emph{Voronoi cell} of $p \in P$ is:
\begin{equation}
V(p) = \{ x\in \mathcal{X} \ | \ \forall q \in P \  d(x,q) \geq d(x,p) \}.
\end{equation}
The \emph{Delaunay triangulation} is the collection of simplices with vertices $\sigma \subseteq P$ such that $\bigcap_{p \in \sigma } V(p) \not = \emptyset$.
\end{definition}

We refer to the $1$-skeleton of the Delaunay triangulation as Delaunay \emph{graph}. In the case of the Euclidean space $\mathcal{X}=\mathbb{R}^n$, the Voronoi cells are convex $n$-dimensional polytopes that intersect at the boundary and cover the ambient space -- see Figure \ref{fig:HyperbolicDelaunay}, left. Moreover, for generic $P$, the Delaunay triangulation is embedded as a collection of non-overlapping simplices that cover the convex hull of $P$.    

\subsection{The Klein-Beltrami Model}\label{sec:Klein_model}
The $n$-dimensional hyperbolic space is, by definition, the unique simply-connected Riemannian manifold with constant curvature equal to $-1$. It admits several isometric models -- see the supplementary material for an overview. As anticipated, we deploy the Klein-Beltrami model since its geodesics are straight lines, which enables to reduce the derivations and computations to the Euclidean case, in particular for computing the hyperbolic Delaunay triangulation. 

The $n$-dimensional Klein-Beltrami model is defined as the Riemannian manifold with ambient space the Euclidean unit ball $\mathbb{K}^{n}= \left\{z \in \mathbb{R}^n \mid\|z\| <1\right\}$,  equipped with the metric tensor:
\begin{equation}
 g(z) = \left( \frac{1}{\| z\|^2 -1 }I_{n} + \frac{1}{(\| z\|^2  -1)^2} z\otimes z \right),
 \end{equation}
where $\otimes$ denotes the tensor product, $I_n$ the identity matrix, and $\langle \cdot, \cdot \rangle$ the Euclidean scalar product. As for any Riemannian manifold, $\mathbb{K}^n$ can be seen as a metric space when equipped with the geodesic distance $d(x,y) = \inf_\gamma \int_{[0,1]} \sqrt{\gamma'(t)^\dagger g(\gamma(t))\gamma'(t)} \ \textnormal{d}t$, where ${\gamma\colon [0,1] \rightarrow \mathbb{K}^n}$ is a smooth curve with $\gamma(0) = x, \gamma(1)=y$. Here, $\gamma'$ denotes the first derivative and $^\dagger$ the transpose. Explicitly, the distance is given by the expression:
\begin{equation}\label{eq:KleinDistance}
d(x,y)= \operatorname{arccosh}\left(\frac{\langle x,y\rangle - 1}{\sqrt{(\| x \|^2  -1)(\| y \|^2  -1)}}\right).
\end{equation}

\section{Method} \label{method}
In this section, we describe HyperDGA -- our proposed method for comparing data representations in a hyperbolic space. An overview of the steps for computing HyperDGA is provided in Algorithm \ref{alg:overview}.

\begin{algorithm}[H]
\caption{HyperDGA Overview}\label{alg:overview}
\KwData{Two datasets $A,B$ represented in a hyperbolic space.}
\KwResult{Similarity score $\textnormal{HyperDGA}(A,B)$.}
\textbf{Step 1:} Convert $A$ and $B$ to $\mathbb{K}^n$ via an isometry. \\ 
\textbf{Step 2:} Compute the hyperbolic Delaunay triangulation of $A \cup B$. \\
\quad\quad\textbf{Step 2.1:} Compute the Euclidean power diagram of $A \cup B$. \\
\quad\quad\textbf{Step 2.2:} Remove the non-Delaunay simplices. \\
\textbf{Step 3:} Compute $\textnormal{HyperDGA}(A,B)$ via Equation \ref{eq:hyperdgamain}. 
\end{algorithm}


\begin{figure*}[!h]
\centering
    \begin{subfigure}[b]{.35\linewidth}
        \centering
        \includegraphics[width=\linewidth]{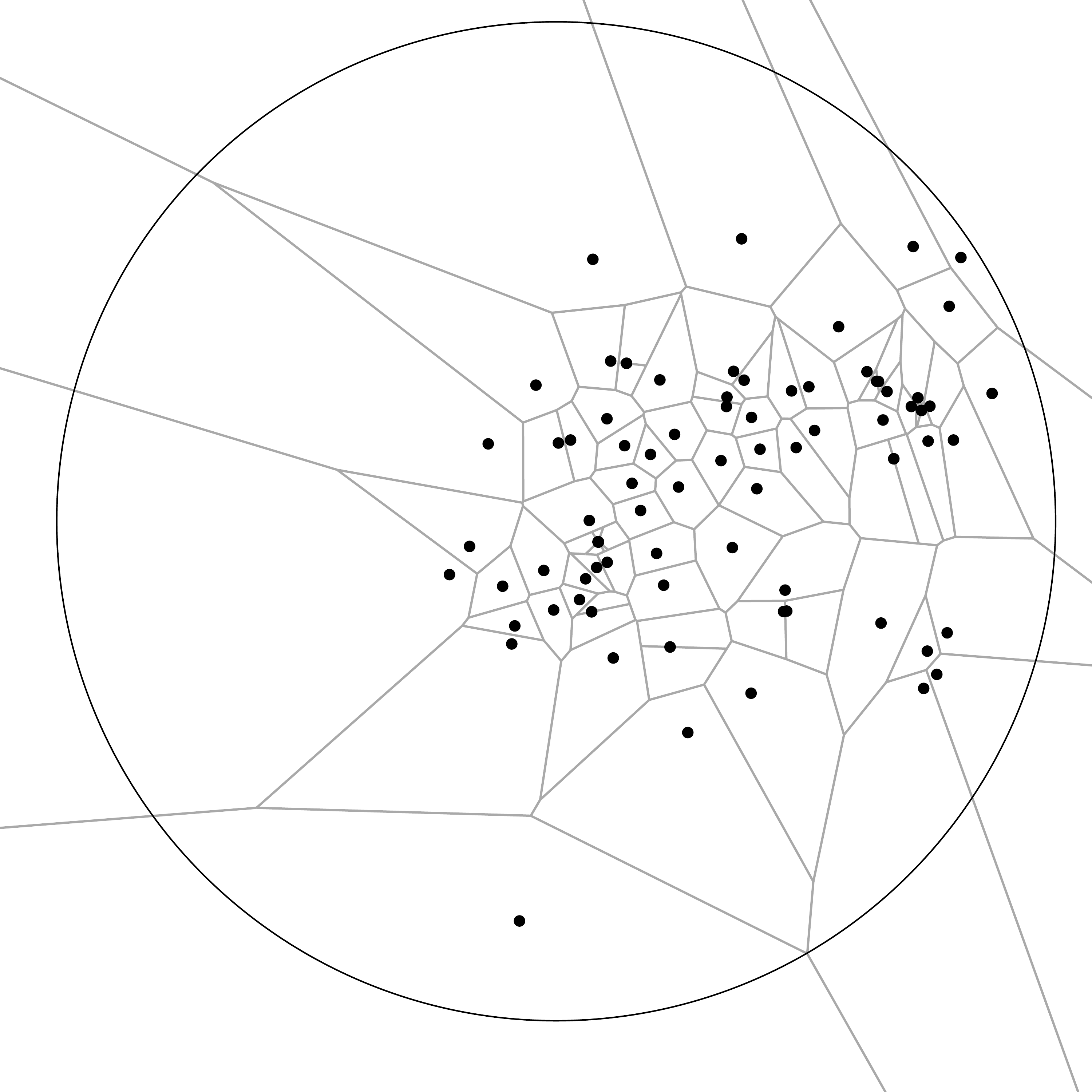}
        \subcaption*{Euclidean power diagram}
    \end{subfigure}
    \hspace{4em}
    \begin{subfigure}[b]{.35\linewidth}
        \centering
 \includegraphics[width=\linewidth]{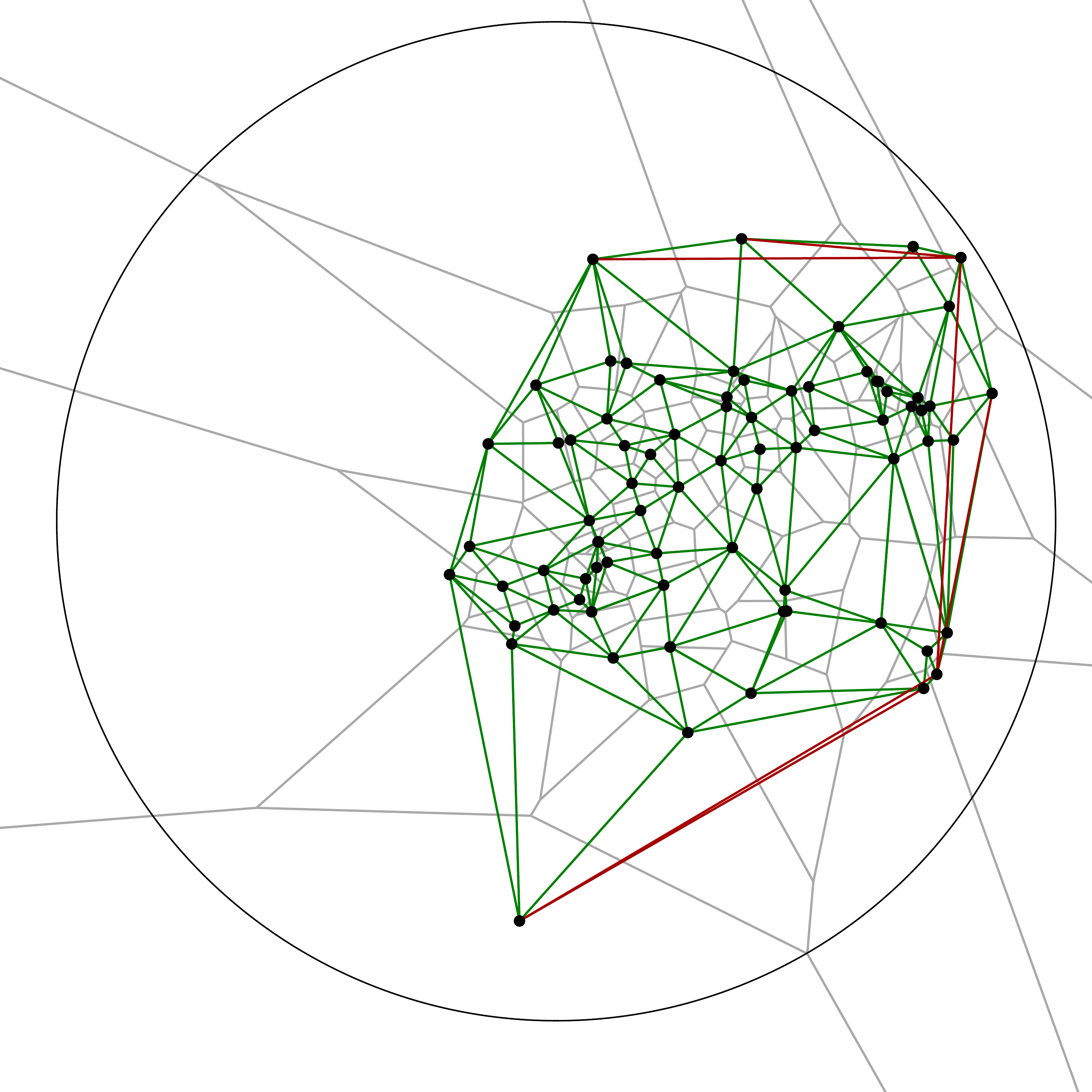}
    \subcaption*{Hyperbolic Delaunay graph}
    \end{subfigure}
    \caption{Depiction of the Euclidean power diagram (gray) equivalent to the hyperbolic Voronoi diagram, the hyperbolic Delaunay graph (green), and the pruned edges (red). Data is obtained from real-life cells (neoblasts 4 \& 7, \cite{Plass2018_planaria_SingleCellData}) via Poincaré embedding \cite{Nickel_2020_SingleCellPoincareEmbedding} and converted to the Klein-Beltrami model.}
    \label{fig:HyperbolicDelaunay}
\end{figure*}

\subsection{Conversion to Klein-Beltrami}
\textbf{Step 1} of Algorithm \ref{alg:overview} consists in converting the datasets $A,B$ from the given model of hyperbolic geometry to the Klein-Beltrami ball model $\mathbb{K}^n$. This is possible since there exist isometric diffeomorphisms between all the hyperbolic models. For example, two popular models in hyperbolic machine learning are the Poincaré ball model $\mathbb{P}^n$ and the Lorentz hyperboloid model $\mathbb{L}^n$ -- see the supplemental material for details. The following maps convert these models to the Klein-Beltrami model:
\begin{itemize} 
\item $z=(z_{0},...,z_{n}) \in \mathbb{L}^n \mapsto \left( \frac{z_{1}}{z_{0}},..., \frac{z_{n}}{z_{0}} \right) \in \mathbb{K}^n$
\item $z \in \mathbb{P}^n \mapsto \frac{2 z}{1+\|z\|^{2}} \in \mathbb{K}^n$.
\end{itemize}

\subsection{Hyperbolic Voronoi Diagram in $\mathbb{K}^n$}\label{sec:HyperbolicVoronoiDiagram}
The goal of \textbf{Step 2} in Algorithm \ref{alg:overview} is to compute the hyperbolic Delaunay triangulation of the union of $A$ and $B$ in the Klein-Beltrami model. As shown by \cite{HyperbolicVoronoiDiagramsMadeEasy2010}, the hyperbolic Voronoi cells in the Klein-Beltrami model correspond to weighted Euclidean Voronoi cells, referred to as \emph{power cells}. We introduce the latter below in a general metric space $\mathcal{X}$.

\begin{definition}\label{def:powercell}
The \emph{power cell} of $p \in P$ with weights $\{ r_p\}_{p \in P} \subseteq \mathbb{R}_{\geq 0}$ is:
\begin{equation}
R(p) = \{ x\in \mathcal{X} \ | \ \forall q \in P \  d_q(x) \geq d_p(x) \},
\end{equation}
where $d_p(x) = d(x,p)^2 - r_p^2$. 
\end{definition}
If all the weights $r_p$ are set to $0$, power cells specialize to Voronoi cells. By leveraging on the non-isometric embedding of the Klein-Beltrami model into the Euclidean space, it is possible to reduce the computation of hyperbolic Voronoi cells to Euclidean power cells with appropriate points and weights. This is performed in \textbf{Step 2.1}, where we compute the Euclidean power diagram of the union of $A$ and $B$ -- see Figure \ref{fig:HyperbolicDelaunay} (left).

\begin{theorem}[\cite{HyperbolicVoronoiDiagramsMadeEasy2010}]\label{thm:hypervoronoi}
Given $P \subseteq \mathbb{K}^n$, there exists an explicit set $S \subseteq \mathbb{R}^n$ and weights $\{ r_s\}_{s \in S}$ such that the hyperbolic Voronoi cells of $P$ correspond to restrictions to $\mathbb{K}^n$ of power cells of $S$. 
\end{theorem}
Explicitly, the set $S$ and weights $\{ r_s\}_{s \in S}$ from Theorem \ref{thm:hypervoronoi} are obtained from points $p \in P$ via:
\begin{equation}
s = \frac{p}{2\sqrt{1- \|p\|^2}}, \hspace{3em} r_s^2=\frac{\|p\|^2}{4(1- \|p\|^2)} - \frac{1}{\sqrt{1- \|p\|^2}}. 
\end{equation}

Power cells, together with their intersections, can be computed in the Euclidean space via standard algorithms from computational geometry, for example, by lifting the points to a hyperboloid and constructing a convex hull \cite{edelsbrunner1985voronoi}. This enables us to compute hyperbolic Voronoi cells, together with hyperbolic Delaunay triangulations, via Theorem~\ref{thm:hypervoronoi}.

However, one subtlety is that power cells might intersect outside the unit ball in $\mathbb{R}^n$. Since these intersections do not define hyperbolic Delaunay simplices, they need to be removed -- see Figure \ref{fig:HyperbolicDelaunay} (right). By duality, the simplices correspond to (potentially unbounded) convex polytopes in $\mathbb{R}^n$. Therefore, it is possible to determine whether the latter intersect $\mathbb{K}^n$ by finding their point of minimum norm, which defines a quadratic programming problem \cite{de2018minimum}. When $n=2$, this reduces to determine whether segments and half-lines intersect the unit circle in the Euclidean plane, which can be addressed by elementary algebraic methods. This is achieved in \textbf{Step 2.2} of Algorithm \ref{alg:overview}.

\subsection{HyperDGA}
\begin{figure*}[!h]
\centering
    \begin{subfigure}[b]{.35\linewidth}
        \centering
        \includegraphics[width=\linewidth]{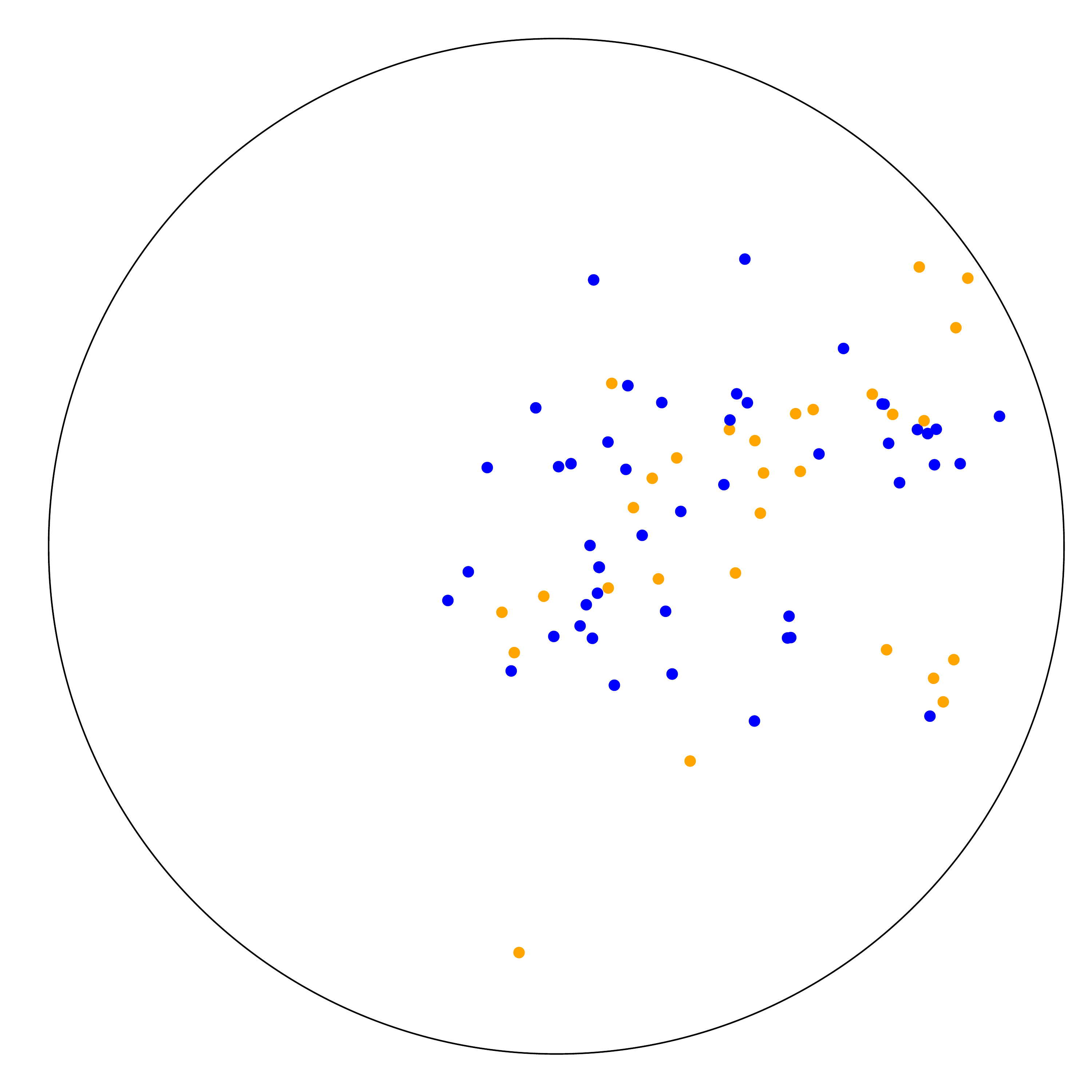}
        \subcaption*{Hyperbolic representation}
    \end{subfigure}
    \hspace{4em}
    \begin{subfigure}[b]{.35\linewidth}
        \centering
 \includegraphics[width=\linewidth]{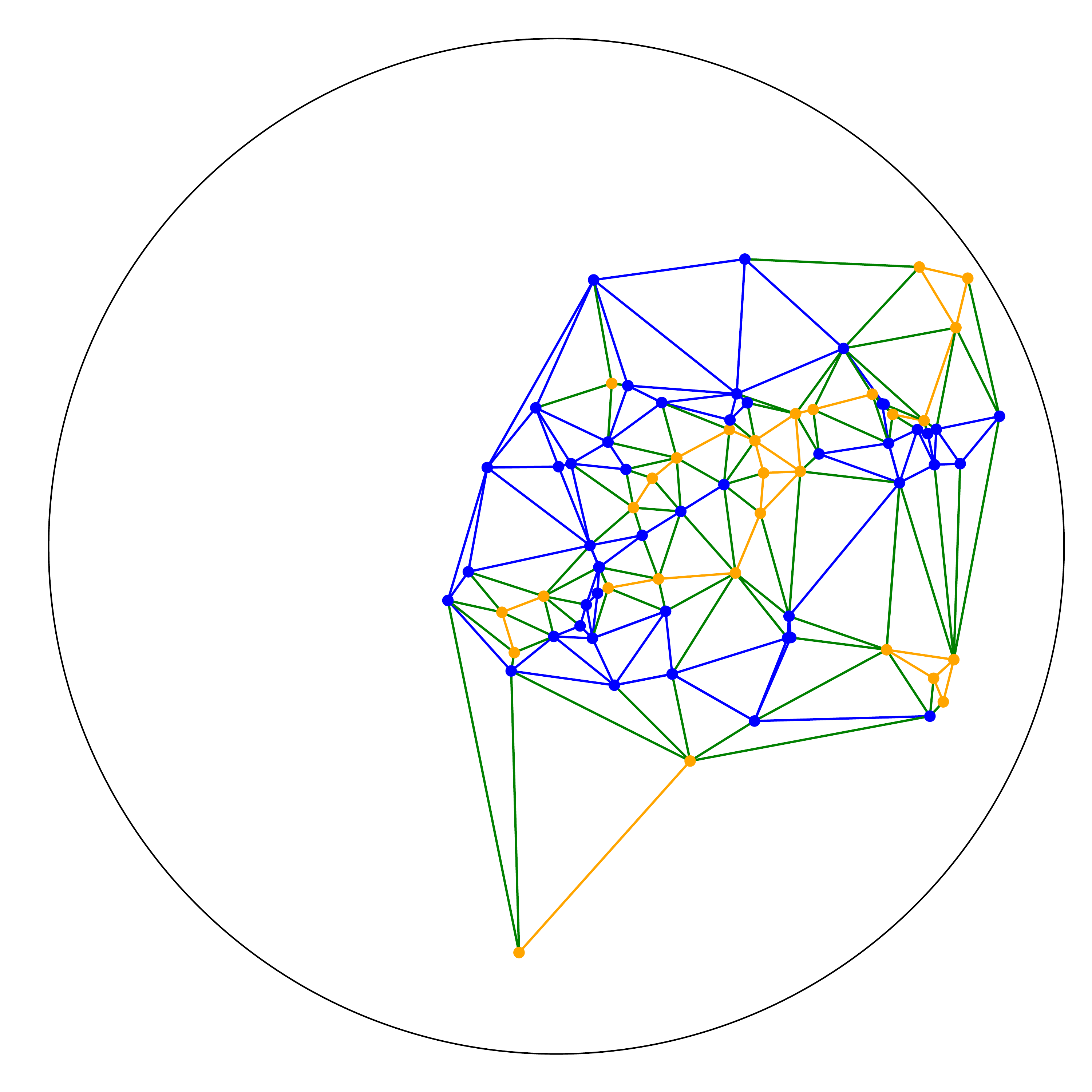}
    \subcaption*{HyperDGA}
    \end{subfigure}
    \caption{Left: Klein-Beltrami hyperbolic representation (via Poincaré embedding \cite{Nickel_2020_SingleCellPoincareEmbedding}) of neoblasts 4 \& 7 \cite{Plass2018_planaria_SingleCellData} (blue and orange). Right: Depiction of their corresponding homogeneous edges (blue and orange) and heterogeneous edges (green).}
    \label{fig:HyperDGA}
\end{figure*}

We now introduce HyperDGA, a similarity score between two sets of datapoints embedded in a hyperbolic space. This is computed in the final \textbf{Step 3} of Algorithm \ref{alg:overview}. Inspired by DCA \cite{Poklukar2022_DCA}, we leverage the Delaunay graph constructed on the union of both datasets. This graph naturally captures the geometric and topological structure of the data by connecting neighboring points via edges whose length adapts to local density changes. The core idea is to measure the geometric alignment of two finite subsets $A, B \subset \mathbb{K}^n$ by counting the proportion of \emph{heterogeneous} edges of the Delaunay graph i.e., the edges connecting points of $A$ and $B$ -- see Figure \ref{fig:HyperDGA} for an illustration. As opposed to heterogeneous edges, the \emph{homogeneous} ones connect pairs of points that both belong to either $A$ or $B$. Intuitively, low HyperDGA scores indicate that the sets $A$ and $B$ are  geometrically aligned. This leads to the following definition. 
\begin{definition}
Let $\mathcal{E}$ be the edges of the Delaunay graph of $A \cup B$ and consider the heterogeneous edges $\widetilde{\mathcal{E}} \subseteq \mathcal{E}$ connecting a point in $A$ to one in $B$. We define: 
\begin{equation}\label{eq:hyperdgamain}
    \textnormal{HyperDGA}(A,B) = 1 - \frac{|\widetilde{\mathcal{E}}|}{|\mathcal{E}|} \in [0,1]. 
\end{equation}
\end{definition}

\section{Experiments}
We provide an empirical investigation of HyperDGA via two experiments on synthetic data and one on real-life data. The goal of the first experiment is to demonstrate the effectiveness of HyperDGA in measuring hyperbolic set distances, and to showcase the capabilities of inferring the noise injected in the original data by analyzing its hyperbolic embedding. The second experiment is aimed at showcasing the relevance of HyperDGA as a tool to evaluate latent representations of a Hyperbolic VAE. Finally, in the last experiment, we validate HyperDGA on real-life biological datasets of varying complexity by showing that the measured hyperbolic distances are coherent with the semantics or domain knowledge. 

We compare HyperDGA to the two baselines given by hyperbolic versions of classical metrics between finite sets. Specifically, we consider the (symmetrized) \emph{Chamfer} and \emph{Wasserstein} distances,  which are defined for $A,B \subset \mathbb{K}^n$ of the same cardinality as: 
\begin{center}
\begin{tabular}{cc}
Wasserstein &   Chamfer 
\vspace{.8em}
\\
$ \displaystyle \min_{\pi \colon A \to B}\sum_{p \in A}d(p,\pi(p))$ & \hspace{3em} $ \displaystyle  \sum_{p \in A} \min_{q \in B} d^2(p,q) + \sum_{q \in B} \min_{p \in A} d^2(p,q)$ 
\end{tabular}
\end{center}
Here, $d$ is the Klein-Beltrami distance (Equation \ref{eq:KleinDistance}) and $\pi$ denotes a bijection between $A$ and $B$ representing a discrete optimal transport.   

All the experiments are performed on a machine with CPU Intel Core i7 and with 16GB of RAM.

\subsection{Synthetic Data with Hyperbolic VAE}
For our two experiments on synthetic data, following \cite{Nagano2019_WrappedNormalDistributionHyperbolicSpaceGradientBasedLearning}, we consider a binary tree $A_{0}$ of depth $d=11$ whose nodes consist of binary sequences representing a simple simulation of protein evolution. By construction, the graph geodesic distance between any pair of nodes in the tree is equal to the Hamming distance between their associated sequences. Furthermore, we consider a perturbed version of $A_{0}$ denoted $A_{\epsilon}$ where we randomly flip the value of each coordinate of each node with probability $\epsilon$. The resulting points in $A_{\epsilon}$ are encoded in a Euclidean space of $2^{d}-1$ dimensions (one dimension for each binary coordinate). We then train a Hyperbolic VAE \cite{Nagano2019_WrappedNormalDistributionHyperbolicSpaceGradientBasedLearning} on a dataset $T$ sampled with $\epsilon=0.1$, in order to represent data in a hyperbolic space. The architecture of the model consists of a Multi Layer Perceptron of depth 3 and 100 hidden variables at each layer for both encoder and decoder, with hyperbolic tangent as the activation function and a latent space of dimension 2. We train the Hyperbolic VAE for 100000 epochs. In what follows, we denote the encoder and the decoder by $E$ and $D$, respectively.

\subsubsection{Experiment 1: Distance Behavior and Noise Inference.}
\begin{figure*}[!h]
         
    \begin{subfigure}[b]{.3\linewidth}
        \centering
        \includegraphics[width=\linewidth]{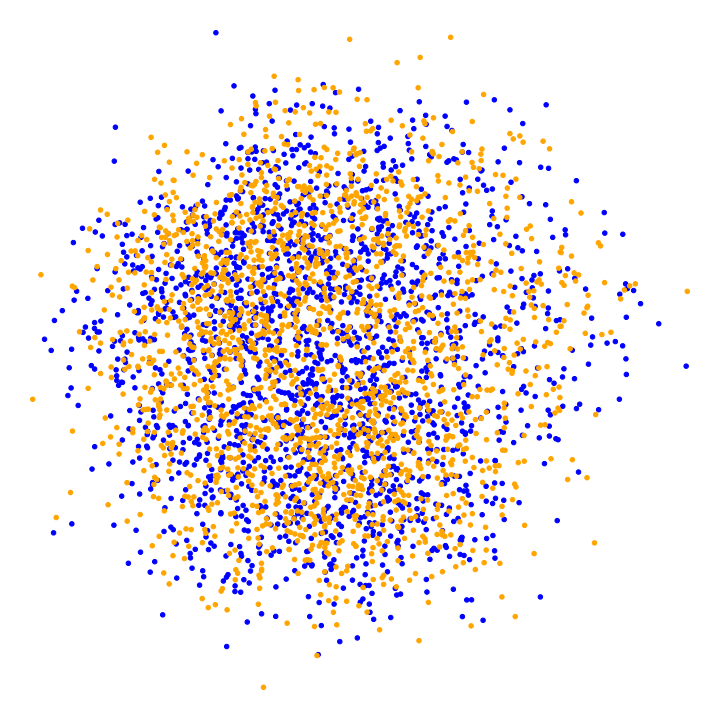}
        \subcaption*{$\epsilon = 0.1$}
    \end{subfigure}
    \hspace{1em}
    \begin{subfigure}[b]{.3\linewidth}
        \centering
 \includegraphics[width=\linewidth]{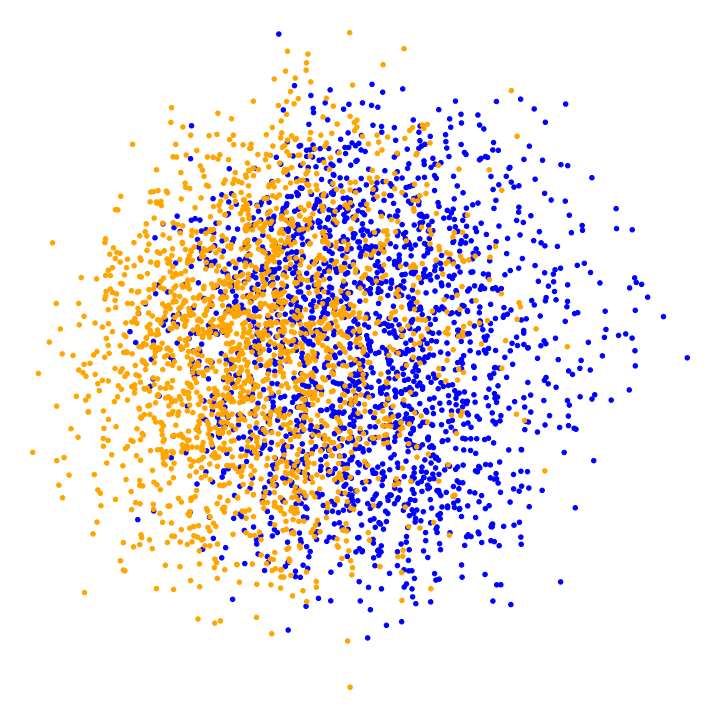}
    \subcaption*{$\epsilon = 0.3$}
    \end{subfigure}
    \hspace{1em}
    \begin{subfigure}[b]{.3\linewidth}
        \centering
        \includegraphics[width=\linewidth]{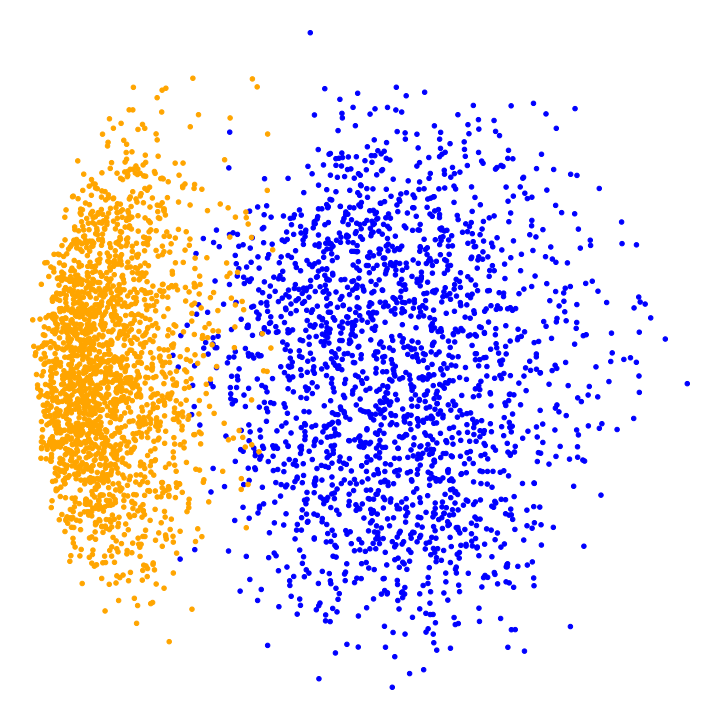}
        \subcaption*{$\epsilon = 0.8$}
    \end{subfigure}
    
    \begin{subfigure}[b]{.3\linewidth}
        \centering
 \includegraphics[width=\linewidth]{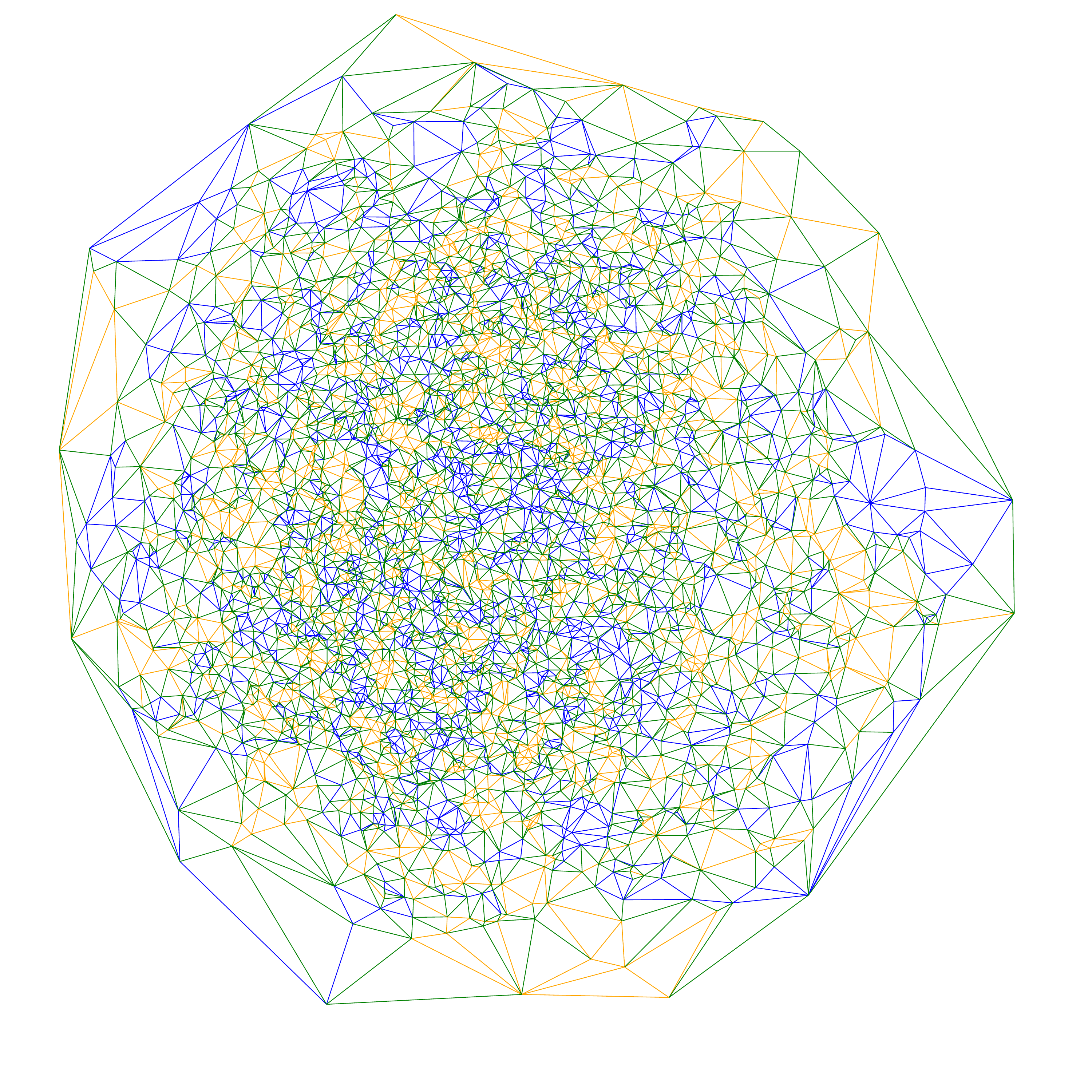}
    \end{subfigure}
    \hspace{1em}
    \begin{subfigure}[b]{.3\linewidth}
        \centering
 \includegraphics[width=\linewidth]{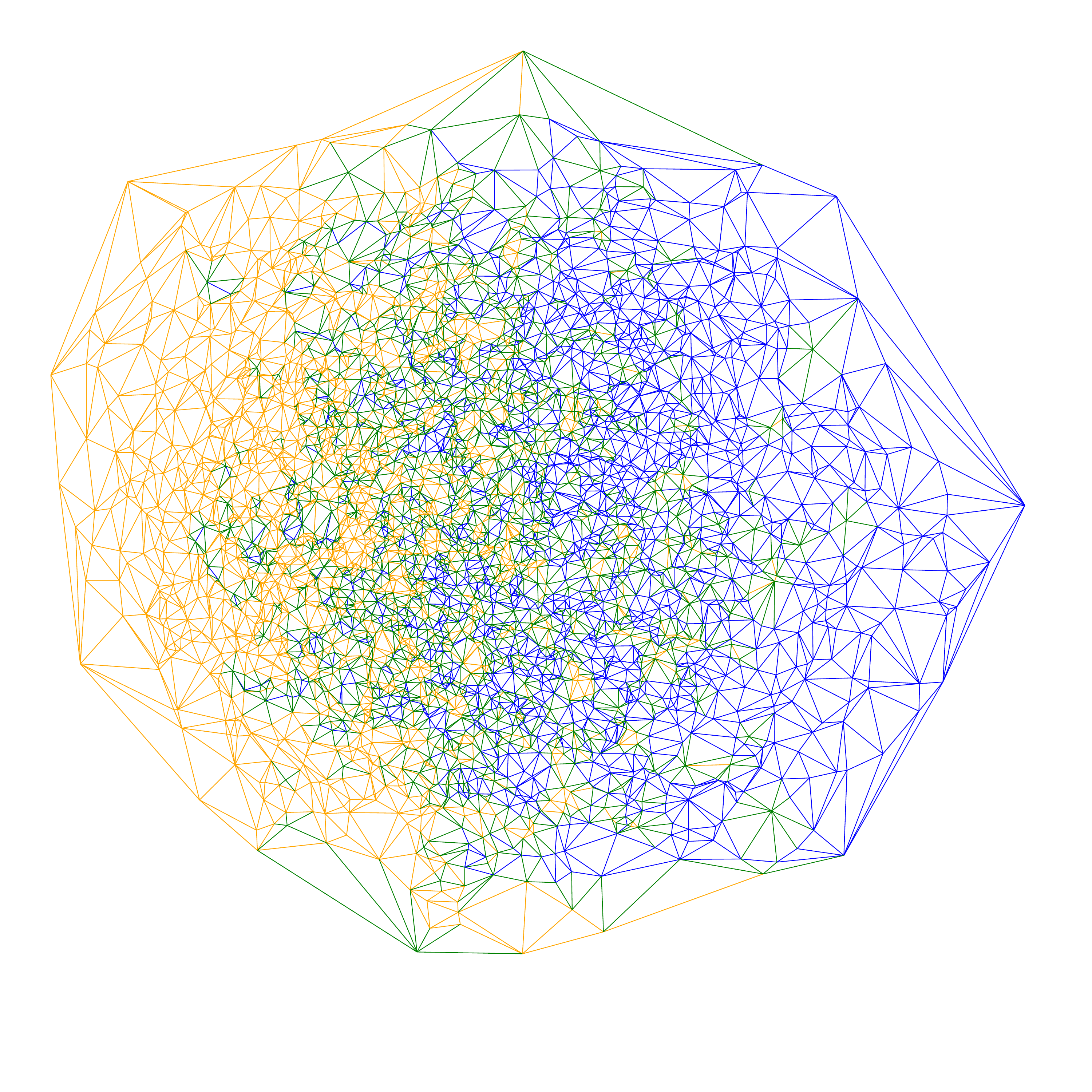}
    \end{subfigure}
    \hspace{1em}
    \begin{subfigure}[b]{.3\linewidth}
        \centering
 \includegraphics[width=\linewidth]{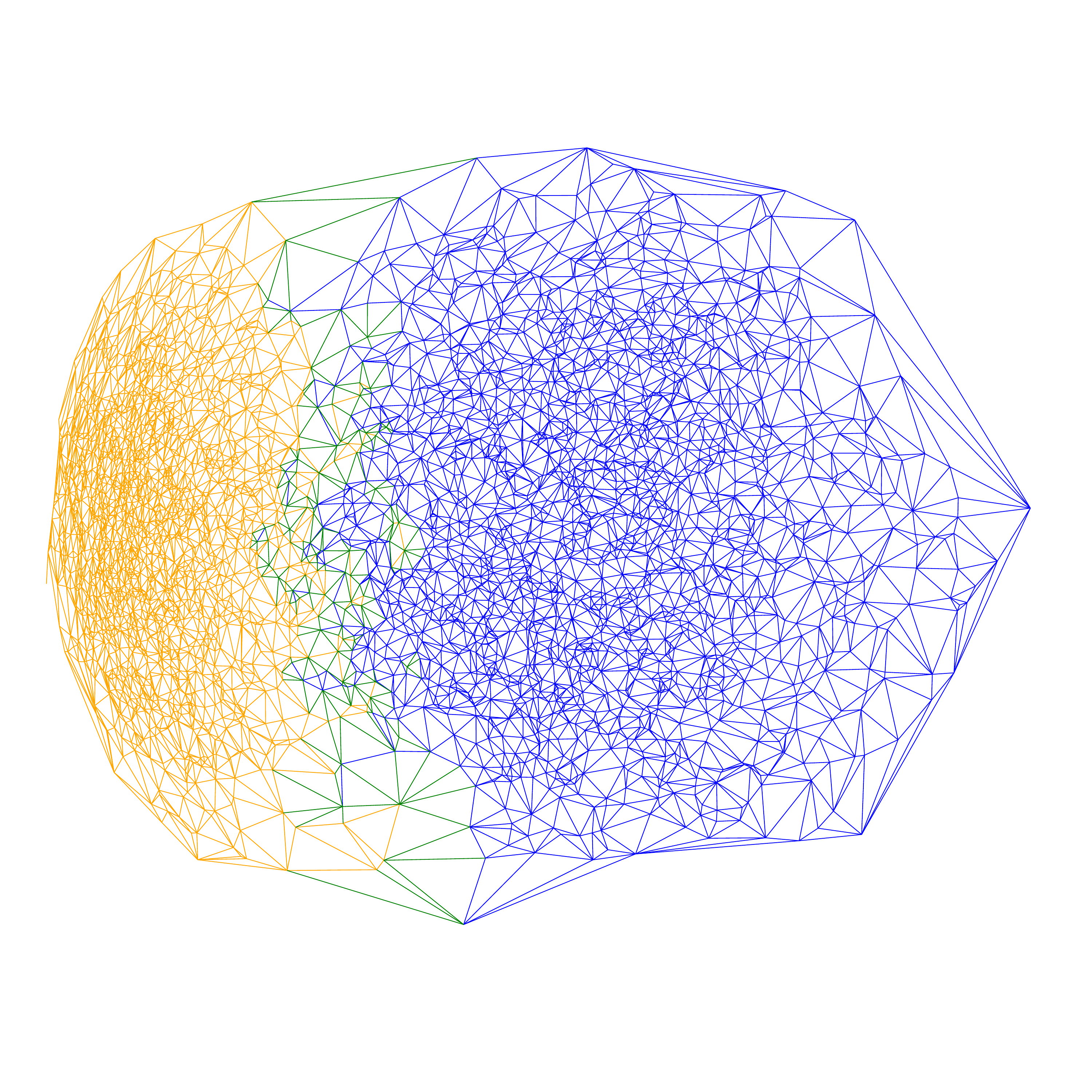}
    \end{subfigure}
    \caption{Top: Poincaré hyperbolic encodings of the training set (blue) and $A_{\epsilon}$ (orange). Bottom: Corresponding Klein-Beltrami visualizations of HyperDGA, with homogeneous edges (blue \& orange) and heterogeneous ones (green).}
    \label{fig:experiment1_visualizations}
\end{figure*}

Following the procedure explained above, we construct 10 synthetic datasets in $2047$ dimensions $A_\epsilon$, $\epsilon \in \{0.1 , 0.2, ..., 0.9, 0.99 \}$. As $\epsilon$ increases, a larger amount of noise is injected in the binary tree data, resulting in more mutations. This can be visualized in Figure \ref{fig:experiment1_visualizations} (top), where the hyperbolic encoding of $A_{\epsilon}$ shifts visibly. Our goal is comparing the encoding $E(T)$ of the training set with the encoding $E(A_\epsilon)$ of the perturbed data. To this end, we measure their similarity in terms of HyperDGA and of the hyperbolic versions of Chamfer and Wasserstein distances. 

\begin{figure}[!h]
 \centering
 \includegraphics[width=.35\linewidth]{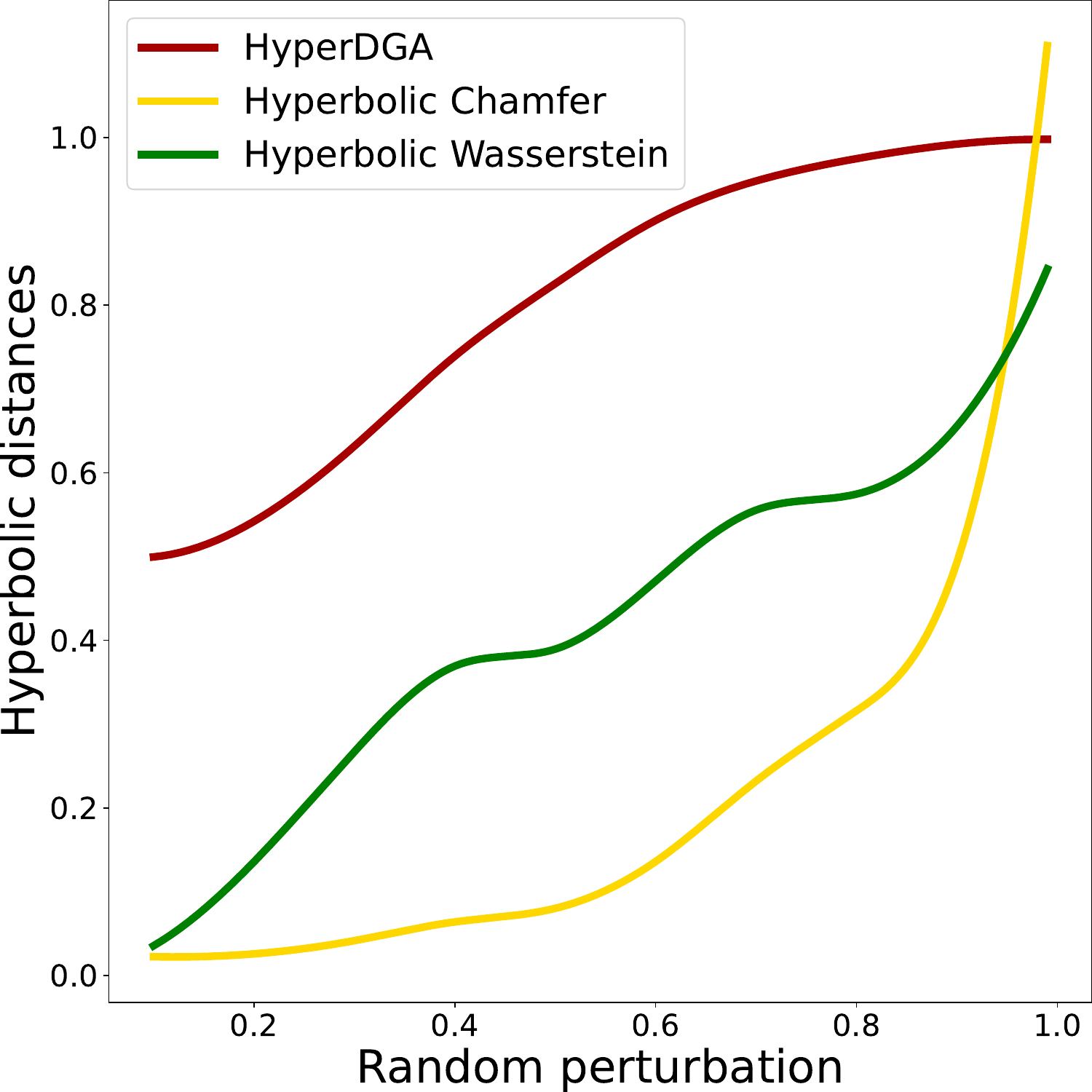} 
\caption{Hyperbolic distances between the encodings of the training set and $A_{\epsilon}$ in function of the random perturbation $\epsilon$.}
\label{fig:experiment1_plots}
\end{figure}

As evident from Figure \ref{fig:experiment1_plots}, $\textnormal{HyperDGA}(E(T), E(A_{\epsilon}))$ is strictly monotonic with respect to $\epsilon$, which is coherent with the increase of the amount of noise in the data. The decrease of the number of heterogeneous edges of the hyperbolic Delaunay graph is visualized in Figure \ref{fig:experiment1_visualizations} (bottom). Moreover, as reported in Table \ref{tab:expperiment1_correlations}, all the considered hyperbolic distances are strongly correlated with the perturbation parameter $\epsilon$. HyperDGA with hyperbolic Wasserstein exhibit the largest correlation. From Figure \ref{fig:experiment1_plots}, it is clear that the noise injected in the original data can be inferred from the latent representation. In other words, in a context of \emph{speciation}, the amount of mutations between two populations can be inferred from their latent representation by measuring distances between sets via HyperDGA or the baselines.

\begin{table}[!h]
    \centering
    \caption{Comparison of the hyperbolic distances in terms of their Pearson correlation with the perturbation $\epsilon$.}

    \begin{tabular}{>{\centering}p{0.3\textwidth}>{\centering}p{0.2\textwidth}>{\centering}p{0.2\textwidth}>{\centering\arraybackslash}p{0.2\textwidth}}
        \toprule
        Hyperbolic distance &  HyperDGA & Chamfer & Wasserstein\\
        \hline
        \hline
        Correl. w/ $\epsilon$ &  0.97 & 0.81 & 0.98\\
        \bottomrule
    \end{tabular}%
    \label{tab:expperiment1_correlations}
\end{table}


\subsubsection{Experiment 2: Evaluation of Latent Representation.}
\begin{figure}[!t]
\centering
    \begin{subfigure}[b]{.2255\linewidth}
        \centering
        \includegraphics[width=\linewidth]{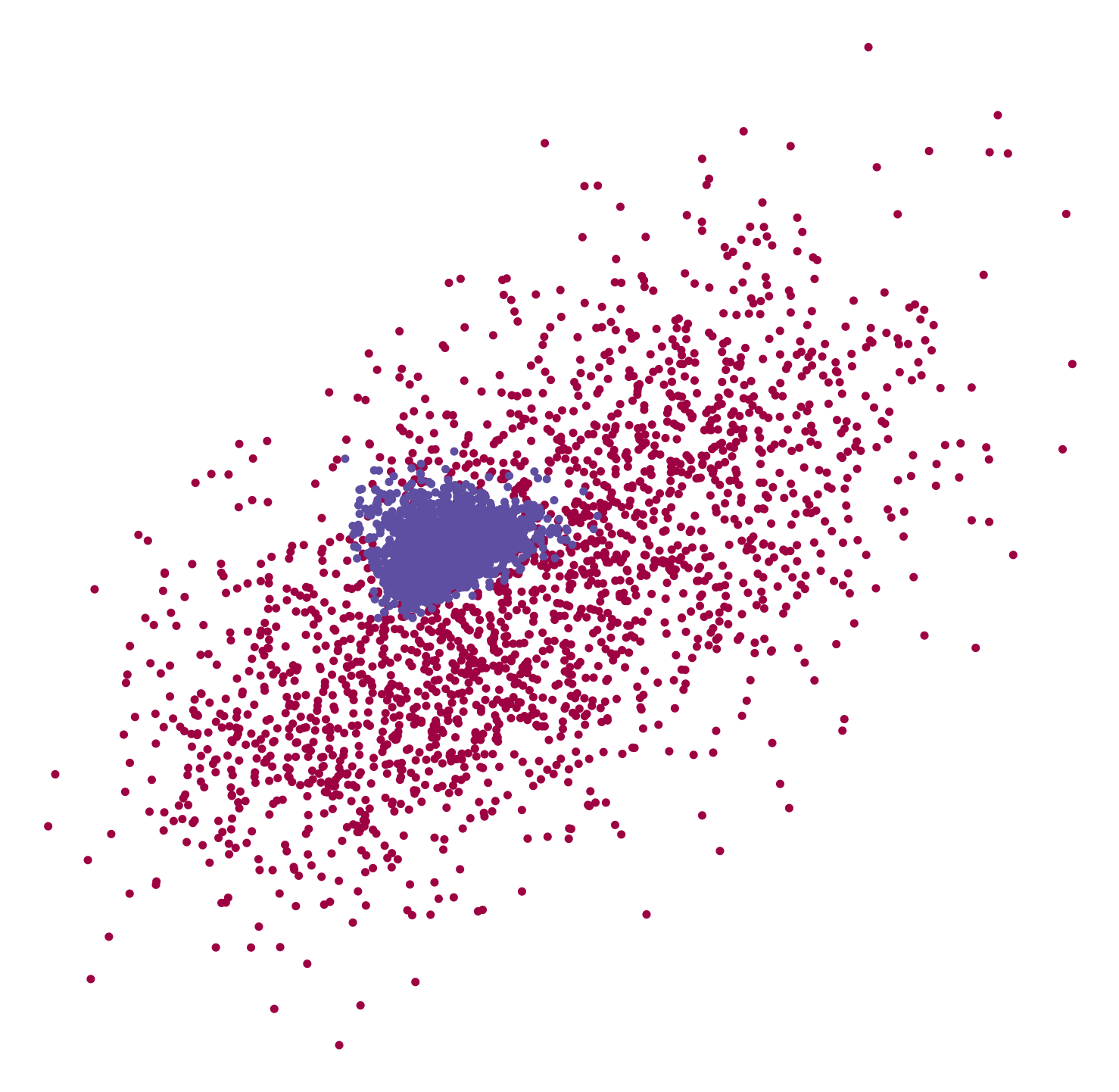}
        \subcaption*{400 epochs}
    \end{subfigure}
    \hspace{1em}
    \begin{subfigure}[b]{.2255\linewidth}
        \centering
 \includegraphics[width=\linewidth]{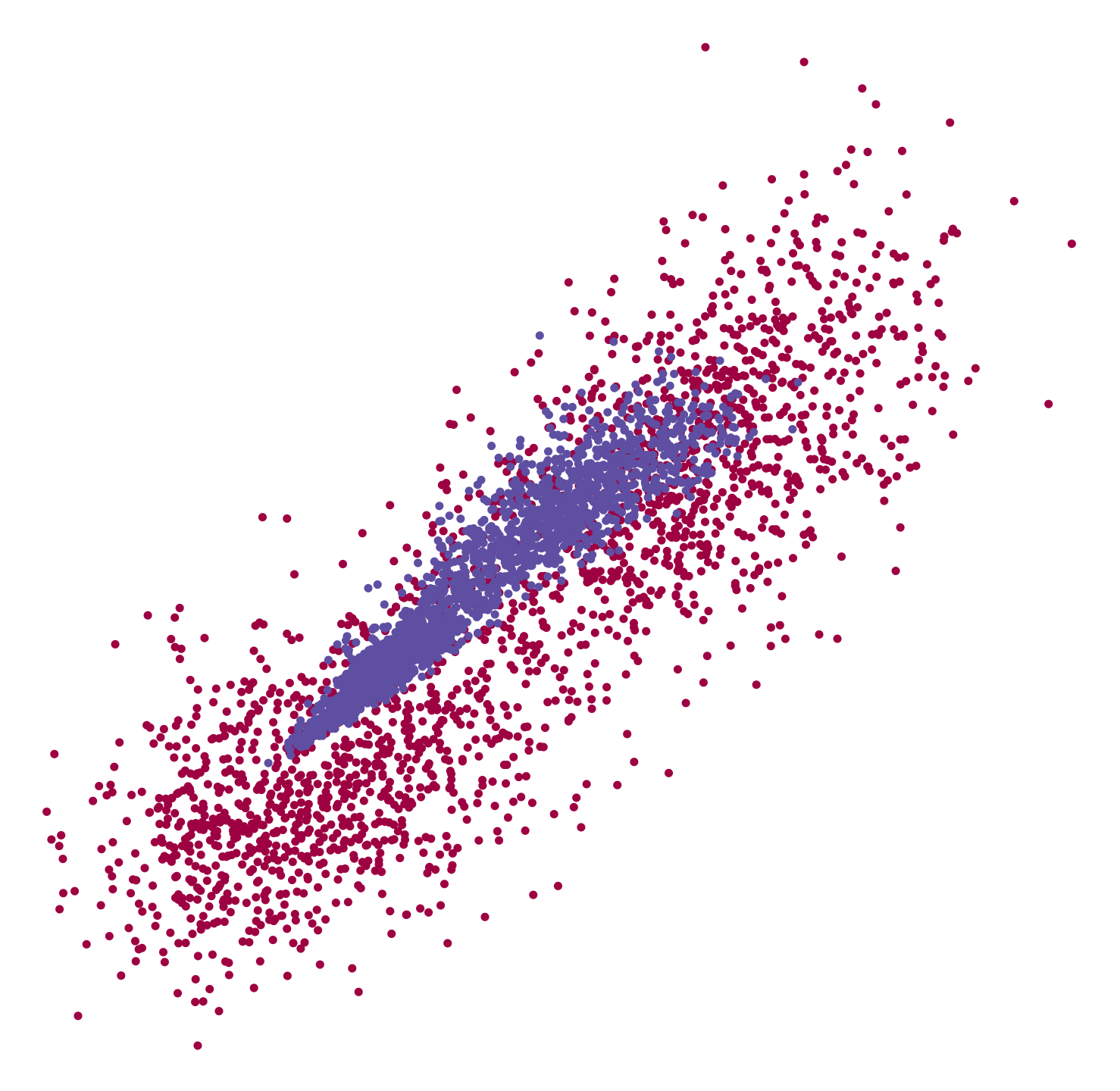}
    \subcaption*{500 epochs}
    \end{subfigure}
    \hspace{1em}
    \begin{subfigure}[b]{.2255\linewidth}
        \centering
        \includegraphics[width=\linewidth]{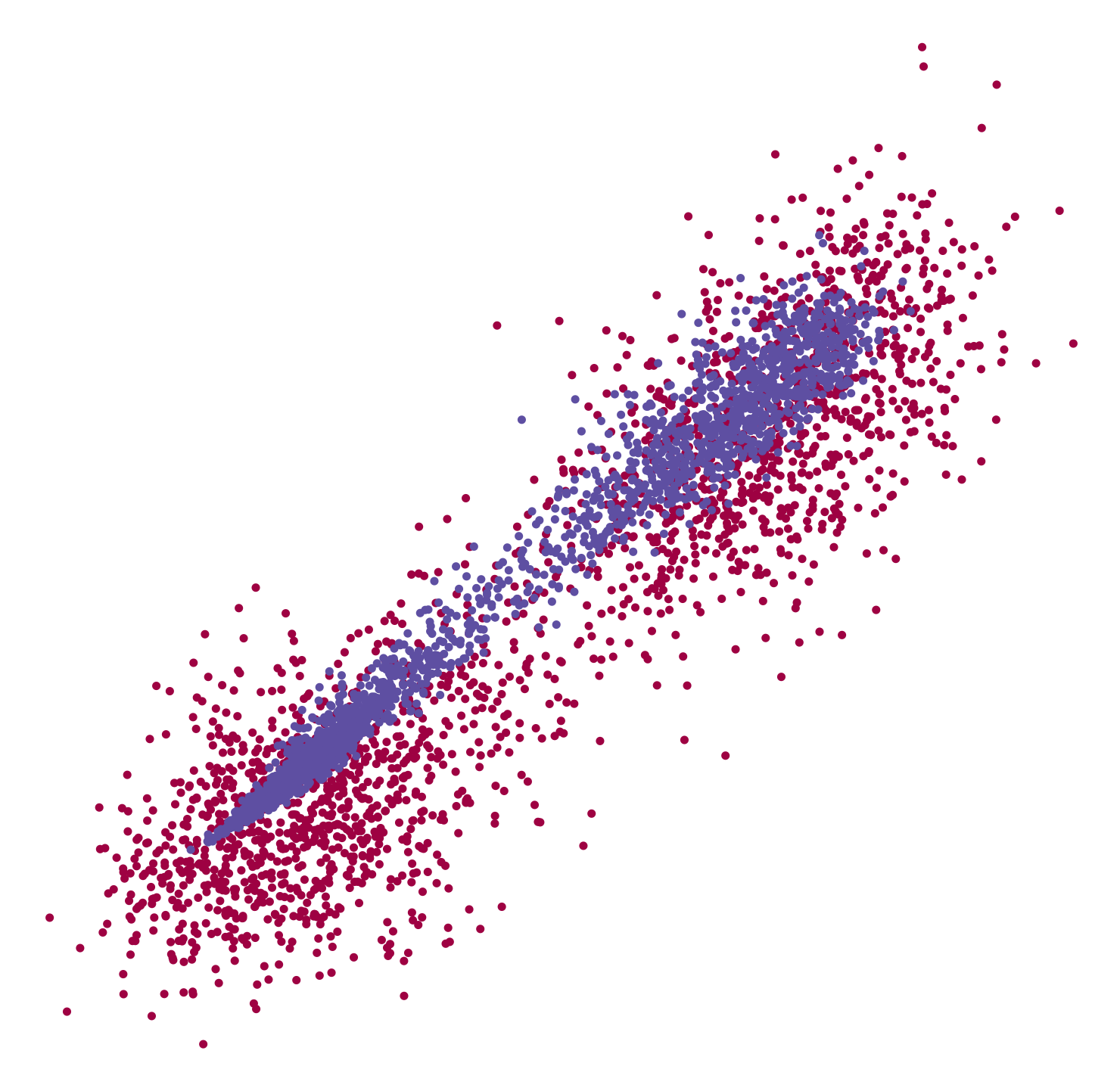}
        \subcaption*{600 epochs}
    \end{subfigure}  
    
    \begin{subfigure}[b]{.2255\linewidth}
        \centering
        \includegraphics[width=\linewidth]{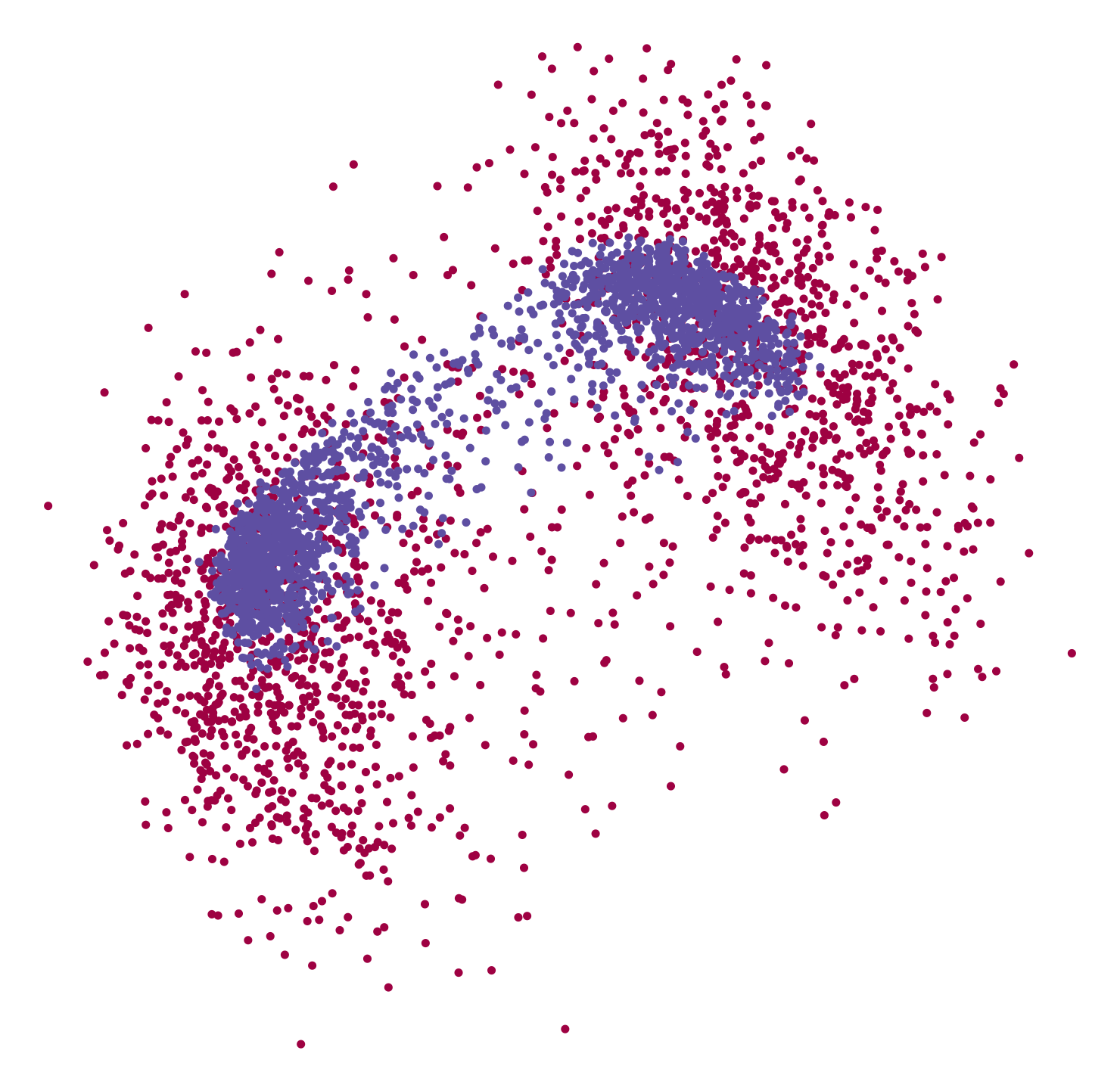}
        \subcaption*{1500 epochs}
    \end{subfigure}
    \hspace{1em}
    \begin{subfigure}[b]{.2255\linewidth}
        \centering
 \includegraphics[width=\linewidth]{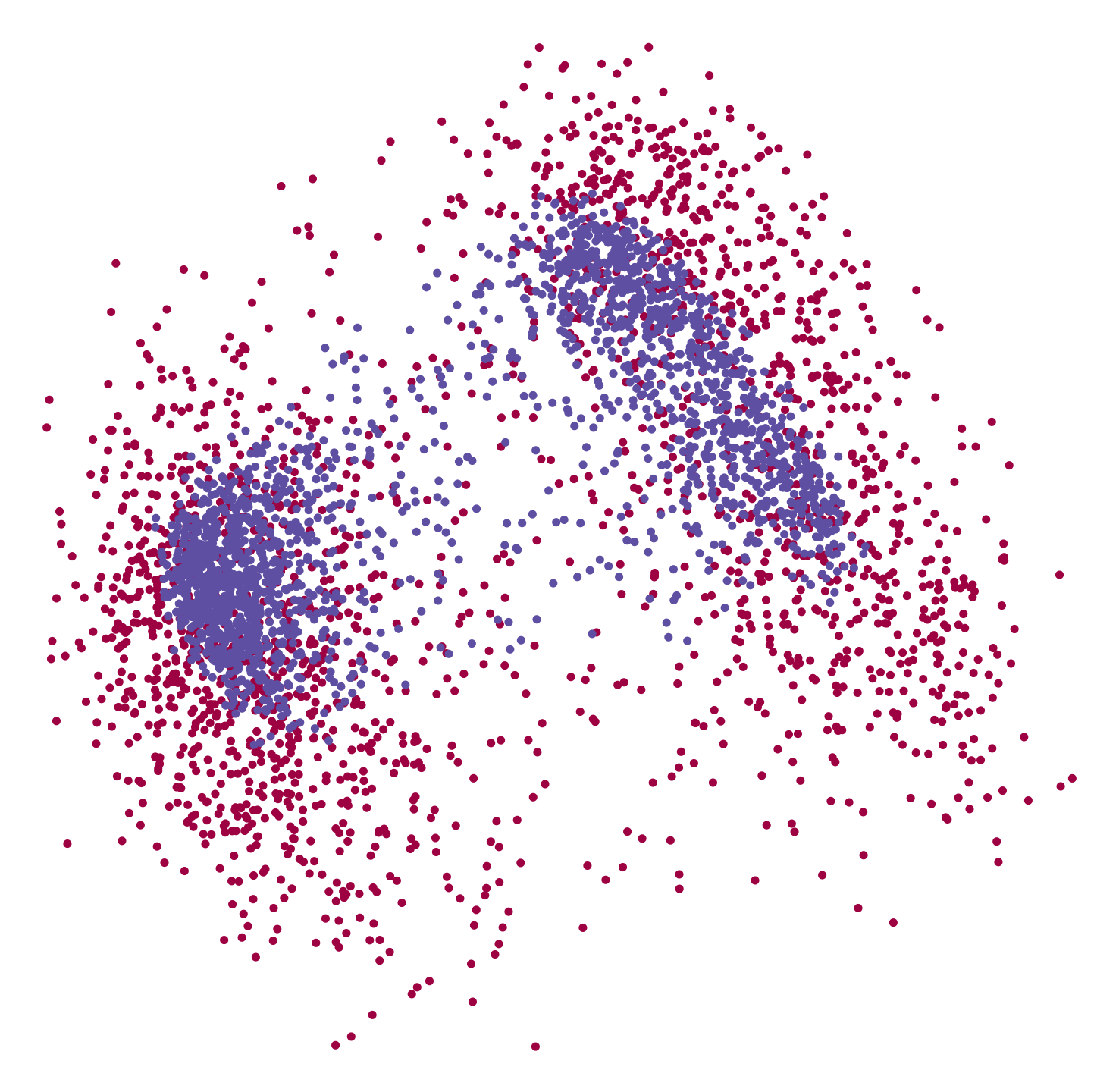}
    \subcaption*{1600 epochs}
    \end{subfigure}
    \hspace{1em}
    \begin{subfigure}[b]{.2255\linewidth}
        \centering
        \includegraphics[width=\linewidth]{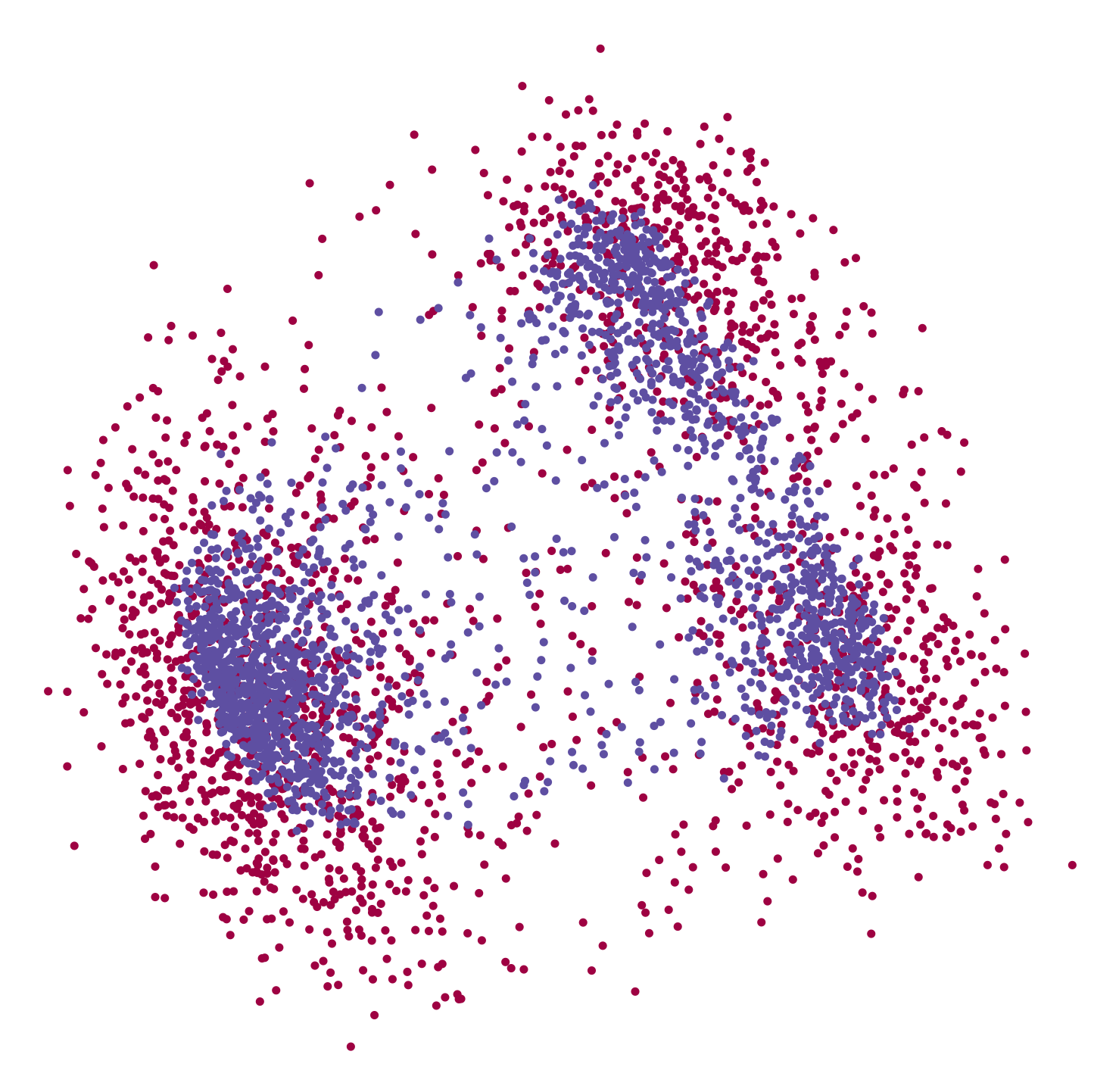}
        \subcaption*{1700 epochs}
    \end{subfigure}
    \caption{Visualization of cluster creation in the encoding of the training set (red) and the encoding of its decoding (purple) while training a Hyperbolic VAE.}
    \label{fig:experiment2_visualizations}
\end{figure}

In this experiment, we compare HyperDGA to the baselines as a method to evaluate the representation inferred by a Hyperbolic VAE. To this end, during training, we measure the given distance between the encoding of the training set $E(T)$ and the encoding of the decoding of its encoding $E(D(E(T)))$, as illustrated in Figure \ref{fig:experiment2_visualizations}. We then investigate the correlation between this quantity and the loss of the VAE, as well as with a supervised performance score. Intuitively, as the training of the Hyperbolic VAE progresses, its latent representation and reconstruction improve, resulting in alignment between the latent distributions we consider. Following \cite{Nagano2019_WrappedNormalDistributionHyperbolicSpaceGradientBasedLearning}, the supervised performance score is given by the absolute difference between the geodesic distances in the binary tree data $A_{0}$ -- computed as the Hamming distance in the input data space -- and the hyperbolic latent distances of the encoding of $A_{0}$. In other words, the latter score evaluates whether the encoder is isometric with respect to the binary tree data. We compute the correlation every 100 epochs of training, up to 1700 epochs.

\begin{figure*}[!b]
         \centering
    \begin{subfigure}[b]{.28\linewidth}
        \centering
        \includegraphics[width=\linewidth]{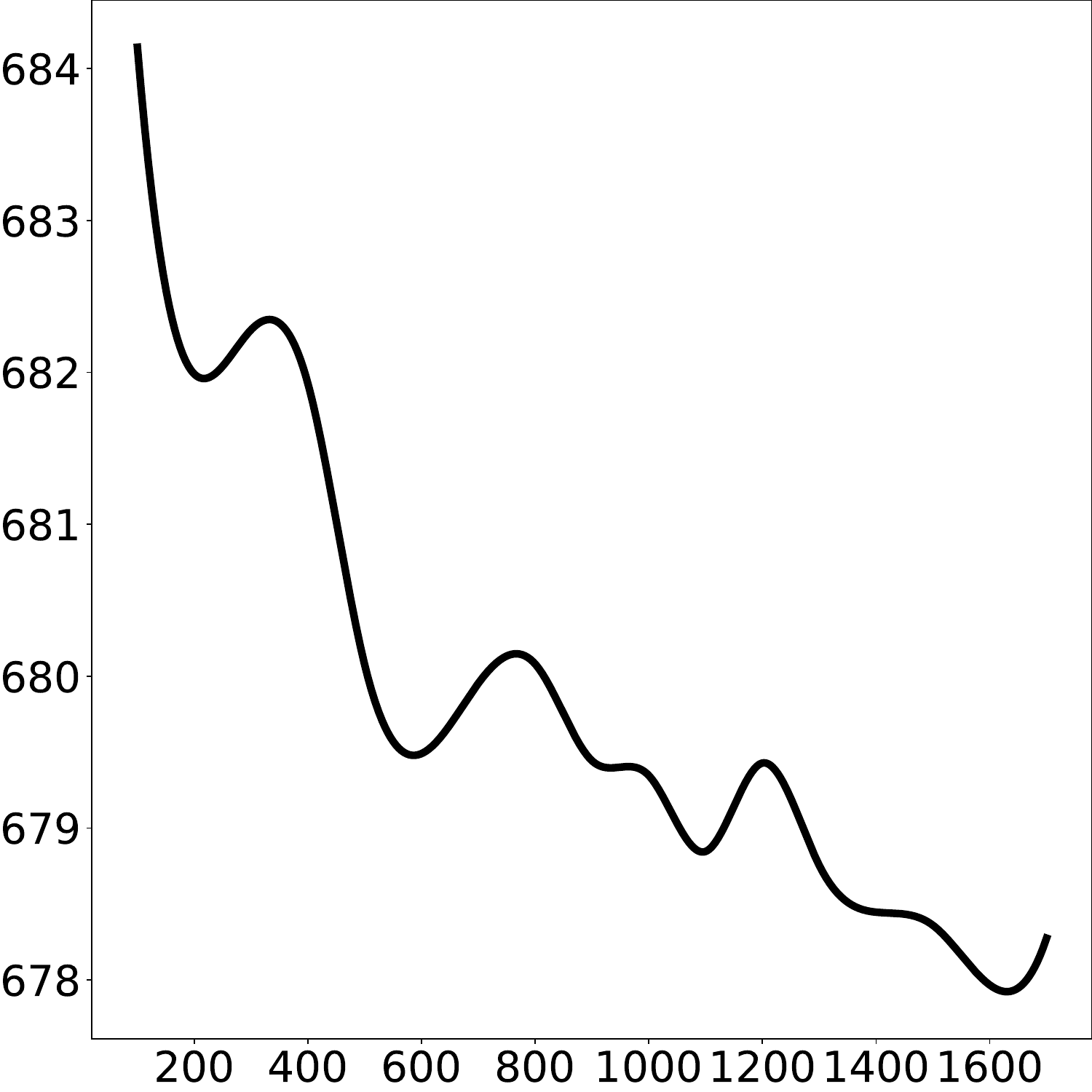}
        \subcaption*{Loss}
    \end{subfigure}
    \hspace{3em}
    \begin{subfigure}[b]{.28\linewidth}
        \centering
 \includegraphics[width=\linewidth]{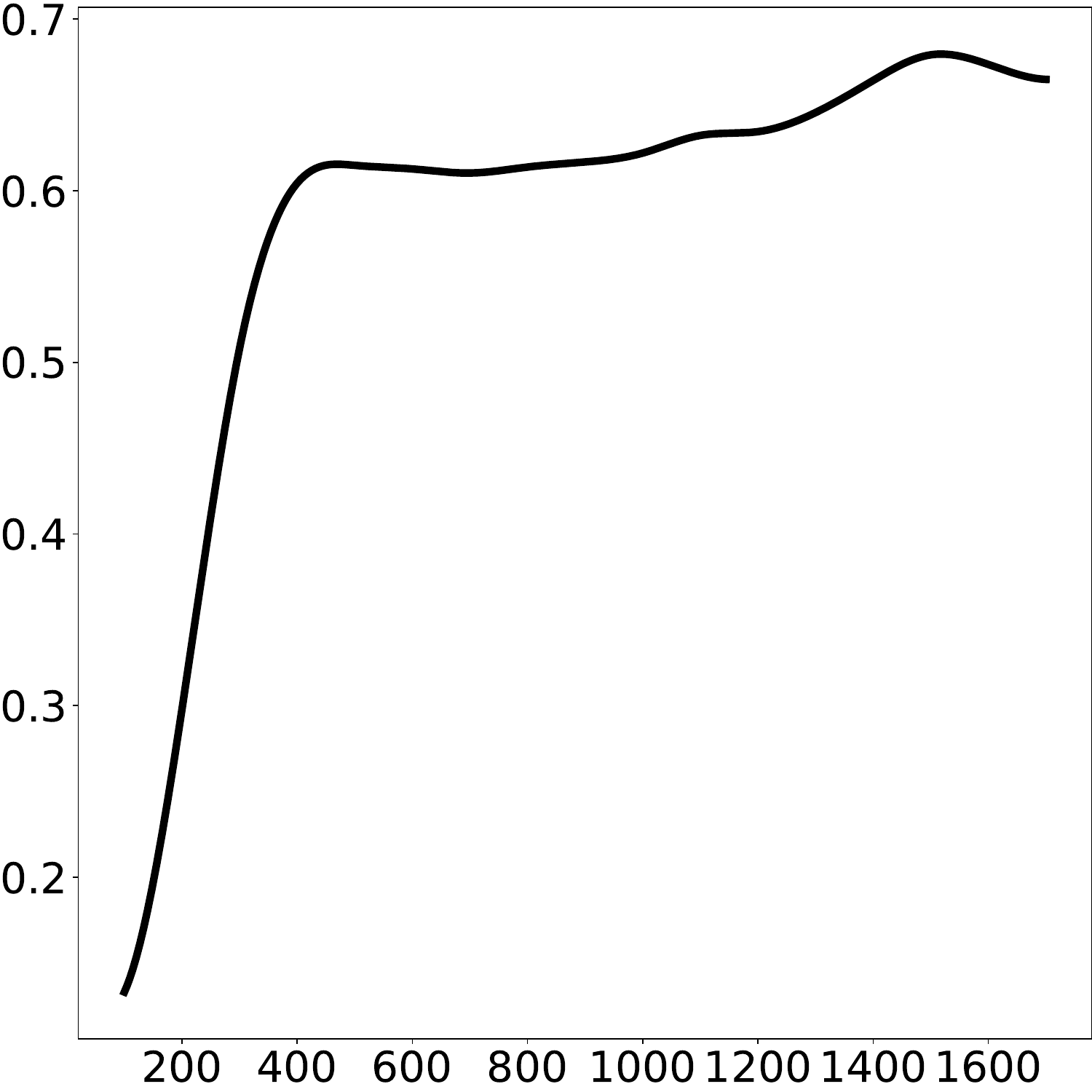}
    \subcaption*{Performance score}
    \end{subfigure}
    
    \vspace{2em}

    \begin{subfigure}[b]{.28\linewidth}
        \centering
        \includegraphics[width=\linewidth]{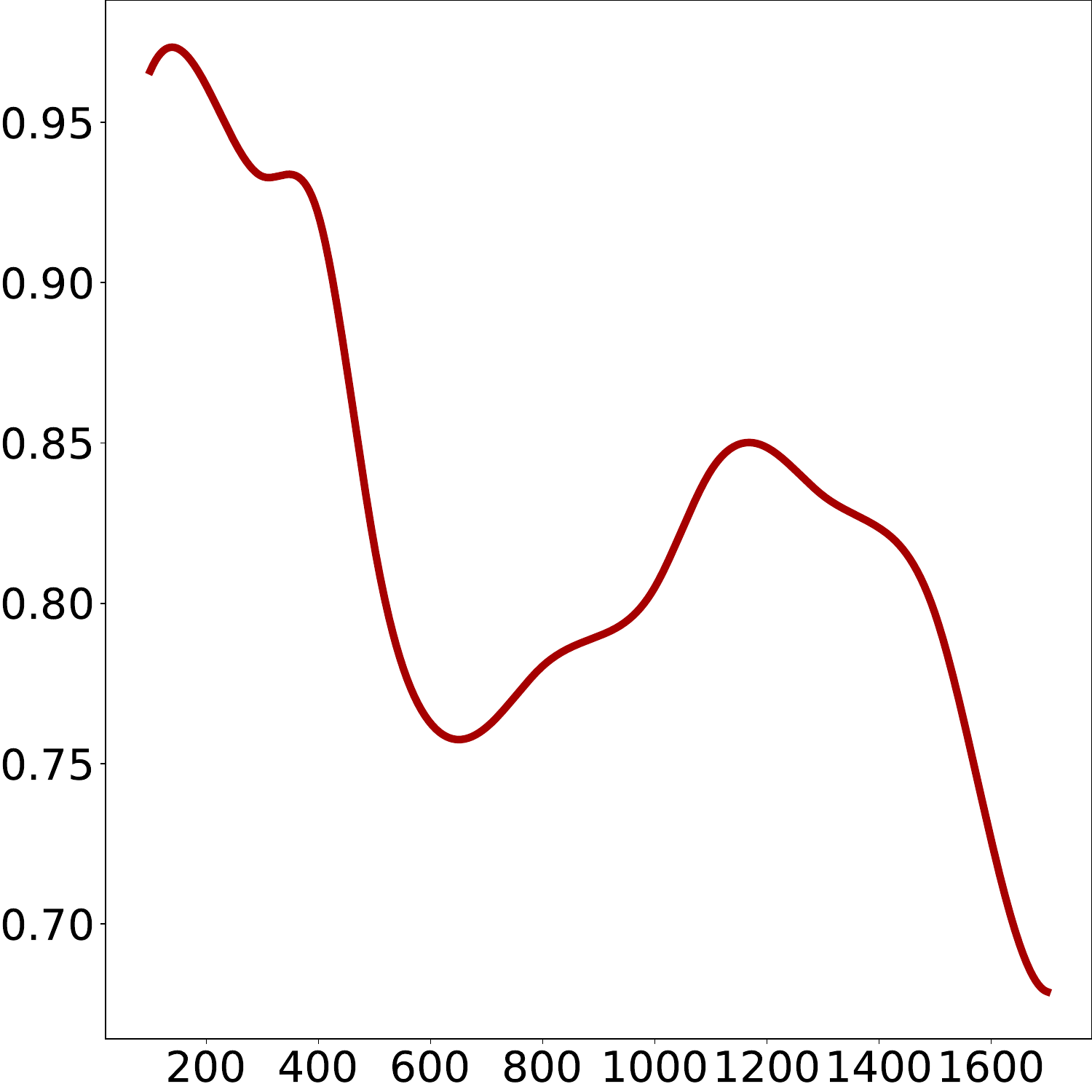}
        \subcaption*{HyperDGA}
    \end{subfigure}
   \hspace{1em}
    \begin{subfigure}[b]{.28\linewidth}
        \centering
        \includegraphics[width=\linewidth]{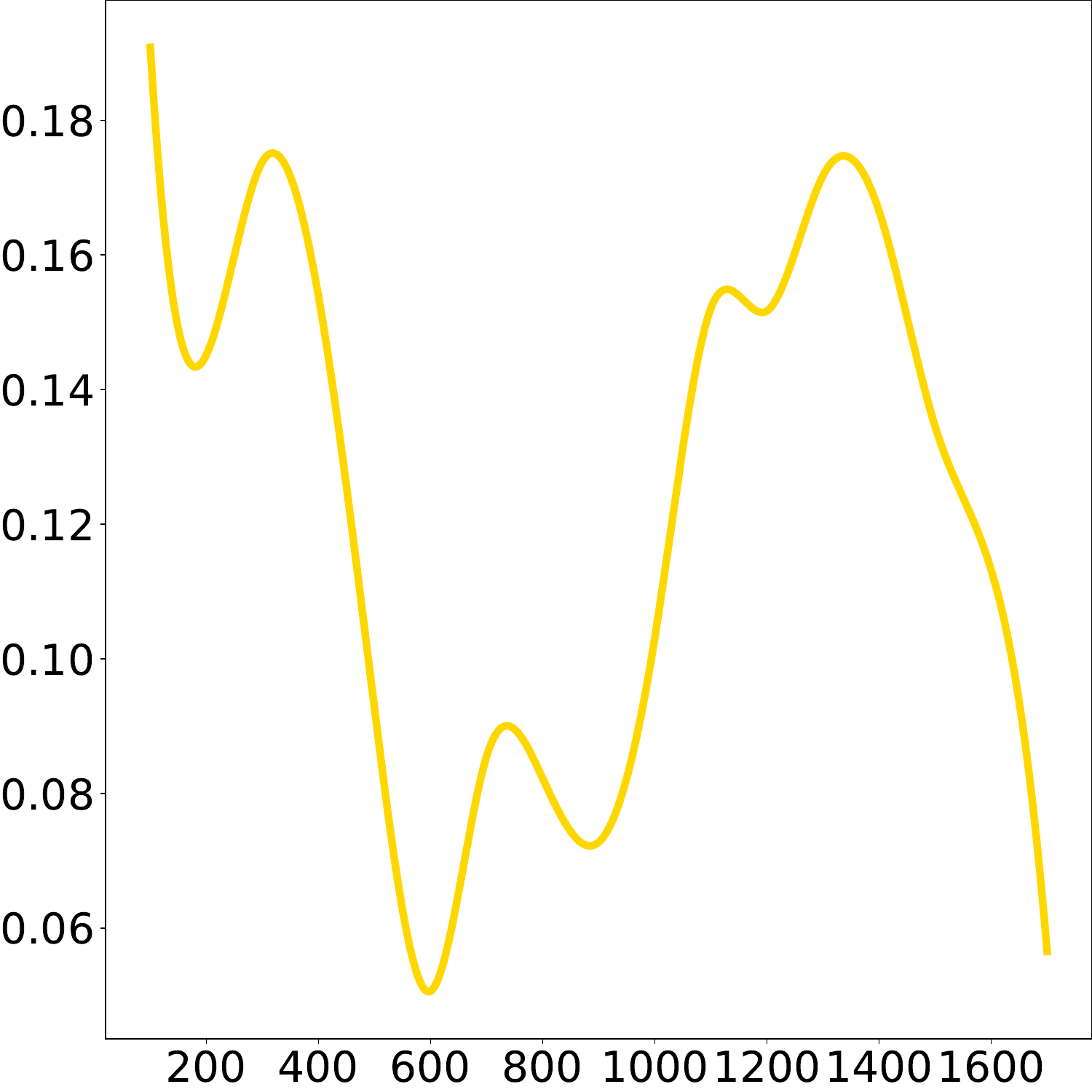}
        \subcaption*{Hyperbolic Chamfer}
    \end{subfigure}
    \hspace{1em}
    \begin{subfigure}[b]{.28\linewidth}
        \centering
        \includegraphics[width=\linewidth]{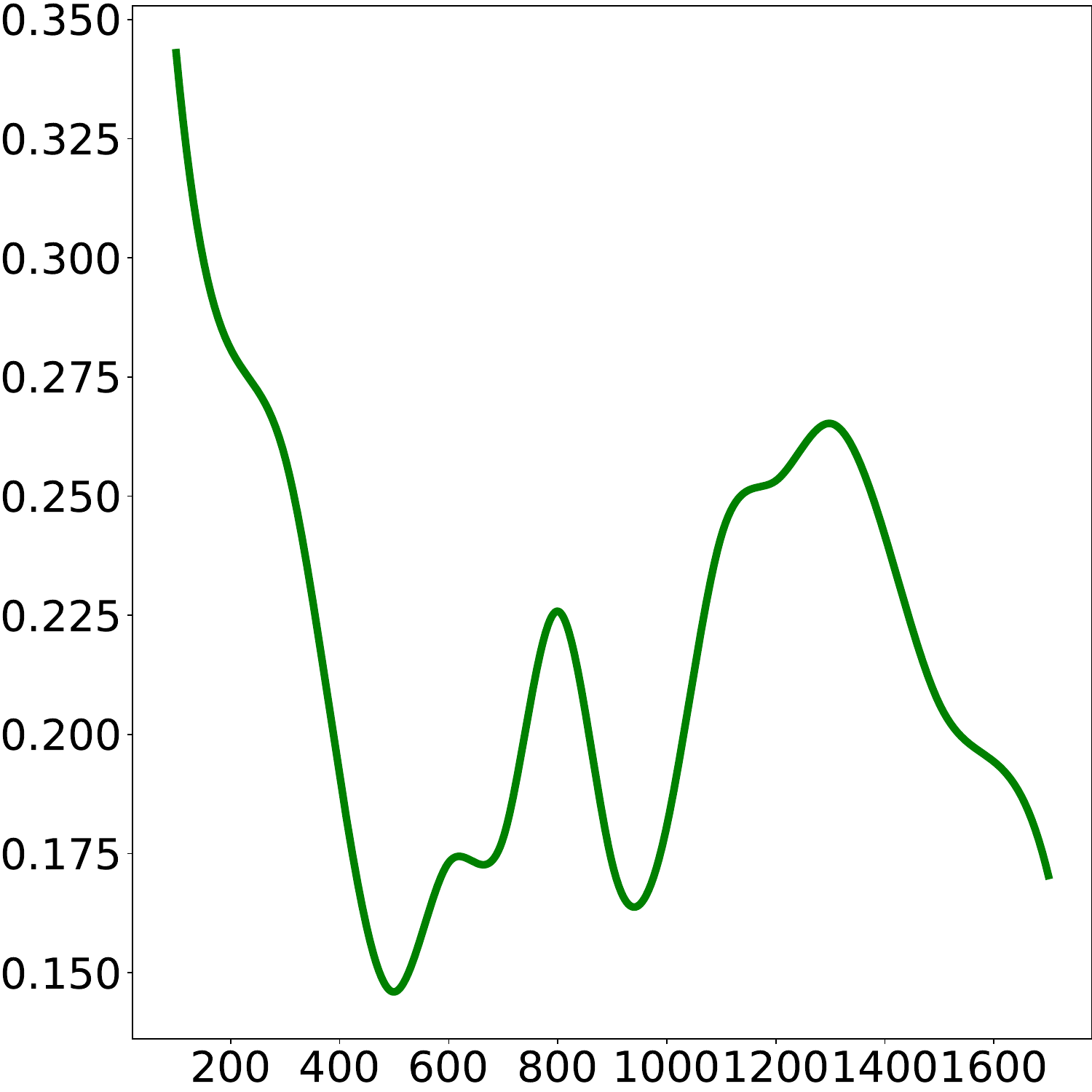}
        \subcaption*{Hyperbolic Wasserstein}
    \end{subfigure}
    \caption{Plots of different metrics in function of the training iteration step, for one seed of the neural network.}
    \label{fig:experiment2_metrics}
\end{figure*}

Table \ref{tab:expperiment2_correlations} reports the means and standard deviations over three random initializations of the VAE. Moreover, Figure \ref{fig:experiment2_metrics} displays plots of the various quantities involved as training progresses, while Figure \ref{fig:experiment2_visualizations} shows visualizations of the hyperbolic latent space. All three methods exhibit a strong Pearson correlation (>0.5) with the loss and with the performance score. Even further, HyperDGA outperforms the baselines in that it exhibits a stronger correlation. Jumps in HyperDGA are associated with the creation of new clusters, as illustrated in Figure \ref{fig:experiment2_visualizations} around epochs 500 (from one to two clusters) and 1600 (from two to three clusters). This means that in addition to the geometric information, HyperDGA takes into account topological changes in the latent representation.
\begin{table}[!h]
    \centering
    \caption{Comparison of the hyperbolic distances in terms of their Pearson correlation with the loss and the performance score.}
    \begin{tabular}{>{\centering}p{0.3\textwidth}>{\centering}p{0.2\textwidth}>{\centering}p{0.2\textwidth}>{\centering\arraybackslash}p{0.2\textwidth}}
        \toprule
        Hyperbolic distance &       HyperDGA &     Chamfer&     Wasserstein\\
        \hline
        \hline
     Correl. w/ Loss &      \textbf{0.76$\pm$0.07}  & 0.52$\pm$0.11 & 0.60$\pm$0.06\\
        Correl. w/ Performance & \textbf{-0.74$\pm$0.07} & -0.58$\pm$0.17 & -0.65$\pm$0.07\\
        \bottomrule
    \end{tabular}
    \label{tab:expperiment2_correlations}
\end{table}

\subsection{Real-Life Biological Data With Poincaré Embedding}
In our last experiment, we validate HyperDGA on hyperbolic embeddings of real-life biological data. In order to embed the data, we rely on the dimensionality reduction technique from \cite{Nickel_2020_SingleCellPoincareEmbedding}. The latter is based on the Poincaré embedding method \cite{Nickel_2017_PoincareEmbedding}, but is specifically designed to visualize single cell data in biology and to uncover hierarchies. We consider three RNA single-cell sequencing datasets to account for different scenarios and complexity in terms of geometric and topological structure in data:
\begin{itemize}
    \item \emph{Olsson}: the mouse myelopoesis data from \cite{Olsson2016_mouse_myelopoesis_SingleCellData} containing 382 cells \footnote{accession code at \url{https://tinyurl.com/olssondata}}.
    \item \emph{Paul}: the mouse myeloid progenitors MARS-seq data from \cite{Paul2015_mouse_myeloid_SingleCellData} containing 2730 cells \footnote{accession code at \url{https://tinyurl.com/pauldata}}.
    \item \emph{Planaria}: the planaria droplet-based single cell transcriptome profiling from \cite{Plass2018_planaria_SingleCellData} containing 21612 cells \footnote{accession code at \url{https://tinyurl.com/plassdata}}.
\end{itemize}
Each dataset consists of several groups of cells and exhibits a hierarchical structure due to the cellular differentiation process where stem cells specialize into, for example, muscle or neuron cells. We expect developmentally similar groups to yield a low score, while dissimilar groups to yield a high score. The pre-processed data are downloaded from \url{https://tinyurl.com/3vw32jad}.


\subsubsection{Experiment 3: Validation on Real-Life Biological Data.}
\begin{figure*}[!h]
    \begin{subfigure}[b]{.32\linewidth}
        \centering
 \includegraphics[width=\linewidth]{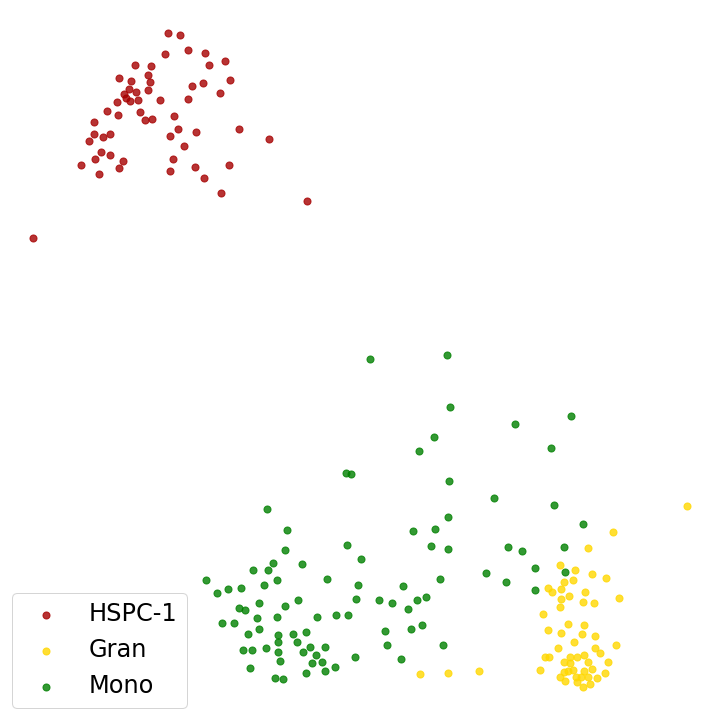}
    \subcaption*{Olsson}
    \end{subfigure}
    \begin{subfigure}[b]{.32\linewidth}
        \centering
        \includegraphics[width=\linewidth]{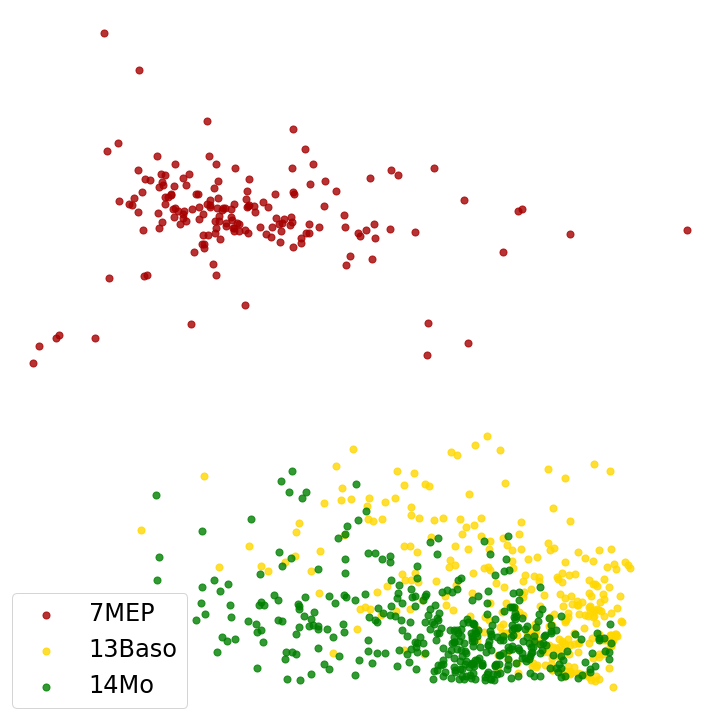}
        \subcaption*{Paul}
    \end{subfigure}
    \centering
    \begin{subfigure}[b]{.32\linewidth}
        \centering
        \includegraphics[width=\linewidth]{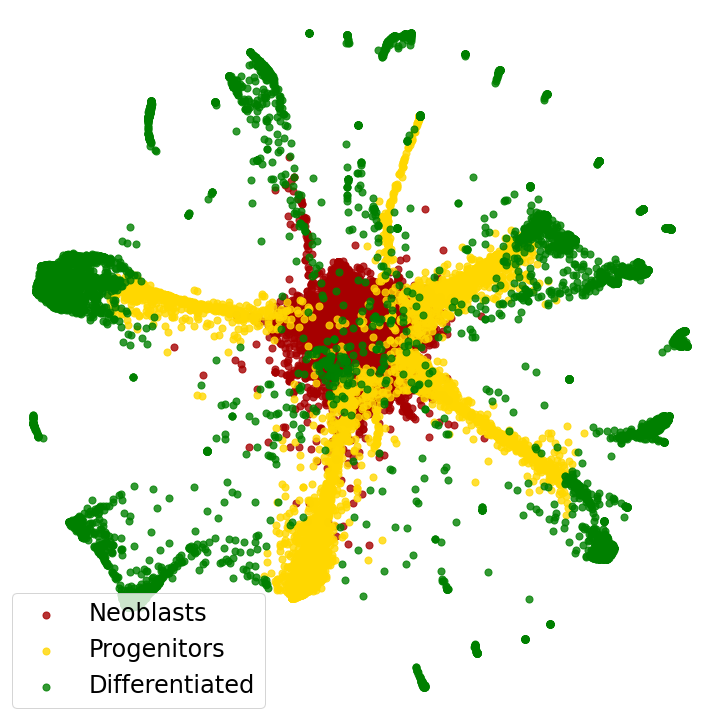}
        \subcaption*{Planaria}
    \end{subfigure}
    \caption{Poincaré embeddings of three classes from three different datasets.}
    \label{fig:experiment3}
\end{figure*}

For each dataset, we measure HyperDGA and the baselines between each pair of three groups of cells and verify that the resulting values are coherent with the domain knowledge. The hyperbolic embeddings of the three groups for each dataset are displayed in Figure \ref{fig:experiment3}. For \emph{Olsson} and \emph{Paul}, we consider two similar classes that are close in the hierarchical tree data: \emph{Mono} and \emph{Gran} for \emph{Olsson}, \emph{13Baso} and \emph{14Mo} for \emph{Paul}. We compare these classes to a third class that lies further in the hierarchy: \emph{HSPC-1} for \emph{Olsson}, \emph{7MEP} for \emph{Paul}. For \emph{Planaria}, we divide the whole dataset in three different groups of cells according to \cite{Plass2018_planaria_SingleCellData}: \emph{Neoblasts}, \emph{Progenitors} and \emph{Differentiated} cells. Given the semantics from biology \cite{Plass2018_planaria_SingleCellData}, we expect the distance between the neoblasts and the differentiated cells to be larger than both the distance between the neoblasts and the progenitors, and the distance between the progenitors and the differentiated cells. 

\begin{table}[!h]
    \centering
    \caption{Hyperbolic distances (HyperDGA, Chamfer, Wasserstein) of three classes for three datasets: \emph{Olsson} (top), \emph{Paul} (middle) and \emph{Planaria} (bottom).}
    \begin{tabular}{>{\centering}p{0.15\textwidth}>{\centering}p{0.25\textwidth}>{\centering\arraybackslash}p{0.25\textwidth}}
            \toprule
         & \emph{Gran} & \emph{HSPC-1} \\
         \hline 
         \hline
        \emph{Mono} & (0.943, 2.6, 0.9) & (0.988, 5.6, 1.3) \\
        \hline
        \emph{Gran} & / & (0.987, 16.7, 2.7) \\
        \bottomrule
    \end{tabular}
    
    \vspace{2em}
    
    \begin{tabular}{>{\centering}p{0.15\textwidth}>{\centering}p{0.25\textwidth}>{\centering\arraybackslash}p{0.25\textwidth}}
    \toprule
         & \emph{14Mo} & \emph{7MEP}\\
        \hline
        \hline
        \emph{13Baso} & (0.713, 0.08, 0.25) & (0.991, 2.70, 0.71)\\
        \hline
        \emph{14Mo} & / & (0.995, 3.18, 0.70)\\
        \bottomrule
    \end{tabular}
    
    \vspace{2em}
    
    \begin{tabular}{>{\centering}p{0.15\textwidth}>{\centering}p{0.25\textwidth}>{\centering\arraybackslash}p{0.25\textwidth}}
    \toprule
         & \emph{Progenitors} & \emph{Differentiated} \\
        \hline
        \hline
        \emph{Neoblasts} & (0.93, 0.07, 0.4) & (0.97, 2.08, 1.2) \\
        \hline
        \emph{Progenitors} & / & (0.96, 1.42, 0.9) \\
        \bottomrule
    \end{tabular}
    \label{tab:exp3}
\end{table}

All the results are reported in Table \ref{tab:exp3}. As evident from the results, HyperDGA and the baselines are consistent with the domain knowledge for all the scenarios among the three datasets. Furthermore, for the embeddings of the neoblasts versus the progenitors from the \emph{Planaria} dataset (see     Figure \ref{fig:experiment3}), we argue that the values of HyperDGA are topologically and geometrically more consistent with the true semantics. Specifically, these two embeddings are visibly different and with significant non-overlapping regions, whereas both Chamfer and Wasserstein give remarkably low distances between them.

\section{Conclusions, Limitations and Future Work}

We have introduced HyperDGA -- a tool for hyperbolic data analysis to complement hyperbolic dimensionality reduction techniques and to evaluate hyperbolic representation learning models. We have demonstrated on synthetic and real-life biological data the relevance of HyperDGA as a method to measure the geometric alignment between two sets represented in a hyperbolic space.

One limitation of HyperDGA is that, by construction, it takes values in rational numbers i.e., $\textnormal{HyperDGA}(A,B) \in \mathbb{Q}$. As a consequence, it is discontinuous and therefore non-differentiable, which is necessary in the context of gradient-based optimization. The latter is potentially useful, since similarity scores can be exploited as objectives for generative modelling and regularization. Therefore, designing continuous and differentiable versions of HyperDGA represents a promising line for future research. 

Another future direction is to investigate practical applications of HyperDGA in hyperbolic machine learning together with domain experts. For example, in the context of biology, similarity scores such as HyperDGA can be deployed to measure alignment between different types of cells, which are commonly represented in a hyperbolic space. This fits within the recent work on hyperbolic embeddings of cells and gene expression \cite{Nickel_2020_SingleCellPoincareEmbedding,Zhou2021HyperbolicGeometryGeneExpression}. In particular, the geometric alignment between tumoral versus healthy cells might provide insights into the evolution of a cancer, leading to a promising biomedical application of HyperDGA.


\section{Acknowledgements}
We wish to thank Yoshihiro Nagano for the technical help, and Mohammad Al-Jaff and Michael Welle for their feedback. This work has been supported by the Swedish Research Council, Knut and Alice Wallenberg Foundation, and the European Research Council through project BIRD (Bimanual Manipulation of Rigid and Deformable Objects).

\newpage 

\bibliographystyle{splncs04}
\bibliography{mybibliography}

\begin{thebibliography}{10}
\providecommand{\url}[1]{\texttt{#1}}
\providecommand{\urlprefix}{URL }
\providecommand{\doi}[1]{https://doi.org/#1}

\bibitem{Beltrami1868_EssayInterpretationNonEuclideanGeometry}
Beltrami, E.: Saggio di interpretazione della geometria Non-Euclidea. s.n. (1868)

\bibitem{Beltrami1868TeoriaFondamentaleSpaziiCurvaturaCostante}
Beltrami, E.: Teoria fondamentale degli spazii di curvatura costante. Annali di Matematica Pura ed Applicata (1867-1897)  \textbf{2},  232--255 (1868)

\bibitem{InternetHyperbolicMapping}
Boguñá, M., Papadopoulos, F., Krioukov, D.: Sustaining the internet with hyperbolic mapping. Nature communications  \textbf{1}, ~62 (09 2010)

\bibitem{book_AlgorithmicGeometry}
Boissonnat, J.D., Yvinec, M.: Algorithmic Geometry. Cambridge University Press (03 1998)

\bibitem{chamberlain2017NeuralEmbeddingHyperbolic}
Chamberlain, B.P., Clough, J., Deisenroth, M.P.: Neural embeddings of graphs in hyperbolic space. arXiv preprint arXiv:1705.10359  (2017)

\bibitem{Chami_2021_HoroPCA}
Chami, I., Gu, A., Nguyen, D.P., Re, C.: Horopca: Hyperbolic dimensionality reduction via horospherical projections. In: Meila, M., Zhang, T. (eds.) Proceedings of the 38th International Conference on Machine Learning. Proceedings of Machine Learning Research, vol.~139, pp. 1419--1429. PMLR (18--24 Jul 2021)

\bibitem{de2018minimum}
De~Loera, J.A., Haddock, J., Rademacher, L.: The minimum euclidean-norm point in a convex polytope: Wolfe's combinatorial algorithm is exponential. In: Proceedings of the 50th Annual ACM SIGACT Symposium on Theory of Computing. pp. 545--553 (2018)

\bibitem{edelsbrunner1985voronoi}
Edelsbrunner, H., Seidel, R.: Voronoi diagrams and arrangements. In: Proceedings of the first annual symposium on Computational geometry. pp. 251--262 (1985)

\bibitem{Fletcher2004PrincipalGeodesicAnalysis}
Fletcher, P., Lu, C., Pizer, S., Joshi, S.: Principal geodesic analysis for the study of nonlinear statistics of shape. Medical Imaging, IEEE Transactions on  \textbf{23},  995 -- 1005 (09 2004)

\bibitem{Guo_2022_CO-SNE}
Guo, Y., Guo, H., Yu, S.X.: Co-sne: Dimensionality reduction and visualization for hyperbolic data. In: Proceedings of the IEEE/CVF Conference on Computer Vision and Pattern Recognition (CVPR). pp. 21--30 (June 2022)

\bibitem{Khrulkov2018_GeometryScore}
Khrulkov, V., Oseledets, I.: Geometry score: A method for comparing generative adversarial networks. In: Dy, J., Krause, A. (eds.) Proceedings of the 35th International Conference on Machine Learning. Proceedings of Machine Learning Research, vol.~80, pp. 2621--2629. PMLR (10--15 Jul 2018)

\bibitem{VAE}
Kingma, D.P., Welling, M.: Auto-encoding variational bayes. In: Bengio, Y., LeCun, Y. (eds.) ICLR (2014)

\bibitem{GeographicRoutingUsingHyperbolic}
Kleinberg, R.: Geographic routing using hyperbolic space. In: IEEE INFOCOM 2007 - 26th IEEE International Conference on Computer Communications. pp. 1902--1909 (2007)

\bibitem{Nickel_2020_SingleCellPoincareEmbedding}
Klimovskaia, A., Lopez-Paz, D., Bottou, L., Nickel, M.: Poincar{\'e} maps for analyzing complex hierarchies in single-cell data. Nature Communications  (2020)

\bibitem{Kynkaanniemi2019_ImprovedPrescisionRecall}
Kynk\"{a}\"{a}nniemi, T., Karras, T., Laine, S., Lehtinen, J., Aila, T.: Improved precision and recall metric for assessing generative models. In: Wallach, H., Larochelle, H., Beygelzimer, A., d\textquotesingle Alch\'{e}-Buc, F., Fox, E., Garnett, R. (eds.) Advances in Neural Information Processing Systems. vol.~32. Curran Associates, Inc. (2019)

\bibitem{Mathieu2019_PoincareVAE}
Mathieu, E., Le~Lan, C., Maddison, C.J., Tomioka, R., Whye~Teh, Y.: Continuous hierarchical representations with poincar\'e variational auto-encoders. In: Advances in Neural Information Processing Systems (2019)

\bibitem{Nagano2019_WrappedNormalDistributionHyperbolicSpaceGradientBasedLearning}
Nagano, Y., Yamaguchi, S., Fujita, Y., Koyama, M.: A wrapped normal distribution on hyperbolic space for gradient-based learning. In: International Conference on Machine Learning (2019)

\bibitem{Nickel_2017_PoincareEmbedding}
Nickel, M., Kiela, D.: Poincar\'{e} embeddings for learning hierarchical representations. In: Guyon, I., Luxburg, U.V., Bengio, S., Wallach, H., Fergus, R., Vishwanathan, S., Garnett, R. (eds.) Advances in Neural Information Processing Systems. vol.~30. Curran Associates, Inc. (2017)

\bibitem{Nickel2018_LorentzEmbedding}
Nickel, M., Kiela, D.: Learning continuous hierarchies in the {L}orentz model of hyperbolic geometry. In: Dy, J., Krause, A. (eds.) Proceedings of the 35th International Conference on Machine Learning. Proceedings of Machine Learning Research, vol.~80, pp. 3779--3788. PMLR (10--15 Jul 2018)

\bibitem{HyperbolicVoronoiDiagramsMadeEasy2010}
Nielsen, F., Nock, R.: Hyperbolic voronoi diagrams made easy. In: 2010 International Conference on Computational Science and Its Applications. pp. 74--80 (2010)

\bibitem{Olsson2016_mouse_myelopoesis_SingleCellData}
Olsson, A., Venkat, M., Chaudhri, V., Aronow, B., Salomonis, N., Singh, H., Grimes, H.L.: Single-cell analysis of mixed-lineage states leading to a binary cell fate choice. Nature  \textbf{537} (08 2016)

\bibitem{Paul2015_mouse_myeloid_SingleCellData}
Paul, F., Arkin, Y., Giladi, A., Jaitin, D., Kenigsberg, E., Keren-Shaul, H., Winter, D., Lara-Astiaso, D., Gury, M., Weiner, A., David, E., Cohen, N., Lauridsen, F., Haas, S., Schlitzer, A., Mildner, A., Ginhoux, F., Jung, S., Trumpp, A., Amit, I.: Transcriptional heterogeneity and lineage commitment in myeloid progenitors. Cell  \textbf{163} (11 2015)

\bibitem{Plass2018_planaria_SingleCellData}
Plass, M., Solana, J., Wolf, F., Ayoub, S., Misios, A., Glažar, P., Obermayer, B., Theis, F., Kocks, C., Rajewsky, N.: Cell type atlas and lineage tree of a whole complex animal by single-cell transcriptomics. Science  \textbf{360},  eaaq1723 (04 2018)

\bibitem{Poklukar2022_DCA}
Poklukar, P., Polianskii, V., Varava, A., Pokorny, F.T., Jensfelt, D.K.: Delaunay component analysis for evaluation of data representations. In: International Conference on Learning Representations (2022)

\bibitem{Poklukar2021_GeomCA}
Poklukar, P., Varava, A., Kragic, D.: Geomca: Geometric evaluation of data representations. In: Meila, M., Zhang, T. (eds.) Proceedings of the 38th International Conference on Machine Learning. Proceedings of Machine Learning Research, vol.~139, pp. 8588--8598. PMLR (18--24 Jul 2021)

\bibitem{VAE_rezende}
Rezende, D.J., Mohamed, S., Wierstra, D.: Stochastic backpropagation and approximate inference in deep generative models. In: Xing, E.P., Jebara, T. (eds.) Proceedings of the 31st International Conference on Machine Learning. Proceedings of Machine Learning Research, vol.~32, pp. 1278--1286. PMLR, Bejing, China (22--24 Jun 2014)

\bibitem{Sala2018_RepresentationTradeoffsHyperbolicEmbeddings}
Sala, F., De~Sa, C., Gu, A., Re, C.: Representation tradeoffs for hyperbolic embeddings. In: Dy, J., Krause, A. (eds.) Proceedings of the 35th International Conference on Machine Learning. Proceedings of Machine Learning Research, vol.~80, pp. 4460--4469. PMLR (10--15 Jul 2018)

\bibitem{Sarkar2012_LowDistortionEmbeddingTreesHyperbolicPlane}
Sarkar, R.: Low distortion delaunay embedding of trees in hyperbolic plane. In: van Kreveld, M., Speckmann, B. (eds.) Graph Drawing. pp. 355--366. Springer Berlin Heidelberg, Berlin, Heidelberg (2012)

\bibitem{Skopek2020_Mixed-curvatureVAE}
Skopek, O., Ganea, O.E., Bécigneul, G.: Mixed-curvature variational autoencoders. In: International Conference on Learning Representations (2020)

\bibitem{Tifrea2019PoincareGlove}
Tifrea, A., Becigneul, G., Ganea, O.E.: Poincaré glove: Hyperbolic word embeddings. In: 7th International Conference on Learning Representations (ICLR) (May 2019)

\bibitem{Zhou2021HyperbolicGeometryGeneExpression}
Zhou, Y., Sharpee, T.O.: Hyperbolic geometry of gene expression. iScience  \textbf{24}(3),  102225 (2021)

\end{thebibliography}
%





\newpage

\section{Appendix}

\subsection{Models of Hyperbolic Geometry}\label{sec:HyperbolicGeometry}

Hyperbolic geometry is a non-Euclidean geometry that has a constant negative Gaussian curvature. Spaces following this geometry are said to be hyperbolic. It is possible to construct several models of hyperbolic geometry, for example the Lorentz hyperboloid and the Poincaré ball model, which are commonly deployed in machine learning to represent hierarchical data. All these models are related by isometric diffeomorphisms.

As emphasized in \cite{Nickel2018_LorentzEmbedding}, the Poincaré ball possesses remarkable visualization capabilities and is intuitively interpretable. On the other hand, the Lorentz hyperboloid model is more convenient for Riemannian optimization and avoids numerical instabilities since its distance function does not involve fractions. We present briefly these two models below. The Poincaré and the Klein-Beltrami model can be seen as map projections of the Lorentz model and were both introduced by Beltrami \cite{Beltrami1868TeoriaFondamentaleSpaziiCurvaturaCostante}.


\subsubsection{Lorentz hyperboloid.}
To begin with, the Lorentzian product -- or, equivalently, the negative Minkowski bilinear form -- between two vectors $z=(z_{0},...,z_{n})$, $z^{\prime}=(z_{0}^{\prime},...,z_{n}^{\prime}) \in \mathbb{R}^{n+1}$ is defined as $\left\langle z,z^{\prime}\right\rangle_{\mathcal{L}}=-z_0 z_0^{\prime}+\sum_{i=1}^n z_i z_i^{\prime}$. The Lorentz hyperboloid model $\mathbb{L}^{n}$ is the set of points $z \in \mathbb{R}^{n+1}$ with $z_{0}>0$ such that its Lorentzian product with itself is equal to $-1$, i.e.; 
\begin{equation}
\mathbb{L}^n=\left\{z \in \mathbb{R}^{n+1} \ | \ \langle z, z\rangle_{\mathcal{L}}= - 1, \ \ z_0>0\right\}.
\end{equation}
The space $\mathbb{L}^n$ is a Riemannian manifold once equipped with the metric tensor given in the standard coordinate system of $\mathbb{R}^{n+1}$ by the diagonal matrix with entries $-1, 1, ..., 1$. The geodesic distance between $z$ and $z^{\prime}$ in $\mathbb{L}^{n}$ is given by 
\begin{equation}
d_{\mathbb{L}^n}(z,z') = \textnormal{arccosh} \left\langle z, z^{\prime}\right\rangle_{\mathcal{L}}.
\end{equation}

\subsubsection{Poincaré ball.}
The $n$-dimensional Poincaré ball model of the hyperbolic space is defined as the Riemannian manifold consisting of the open Euclidean ball $\mathbb{P}^{n} = \left\{z \in \mathbb{R}^n \mid\|z\|^2<1 \right\}$, equipped with the metric tensor given in the standard coordinate system of $\mathbb{R}^n$ by the diagonal matrix with entries equal to $ \frac{4}{(1-\|z\|^{2})^2}$, $z \in \mathbb{P}^n$. The geodesic distance between $z$ and $z^{\prime}$ in $\mathbb{P}^{n}$ is given by 
\begin{equation}
d_{\mathbb{P}^n}(z,z') = \textnormal{arccosh}\left(1 + \frac{2\|z-z'\|^2}{\left(1-\|z\|^2\right)\left(1-\|z'\|^2\right)}\right).
\end{equation}
The geodesics in the Poincaré ball model are portions of circular arcs orthogonal to the boundary of the open ball.

\newpage

\subsection{Additional Real-Life Visualizations}

\begin{figure*}[!h]
    \begin{subfigure}[b]{.32\linewidth}
        \centering
        \includegraphics[width=\linewidth]{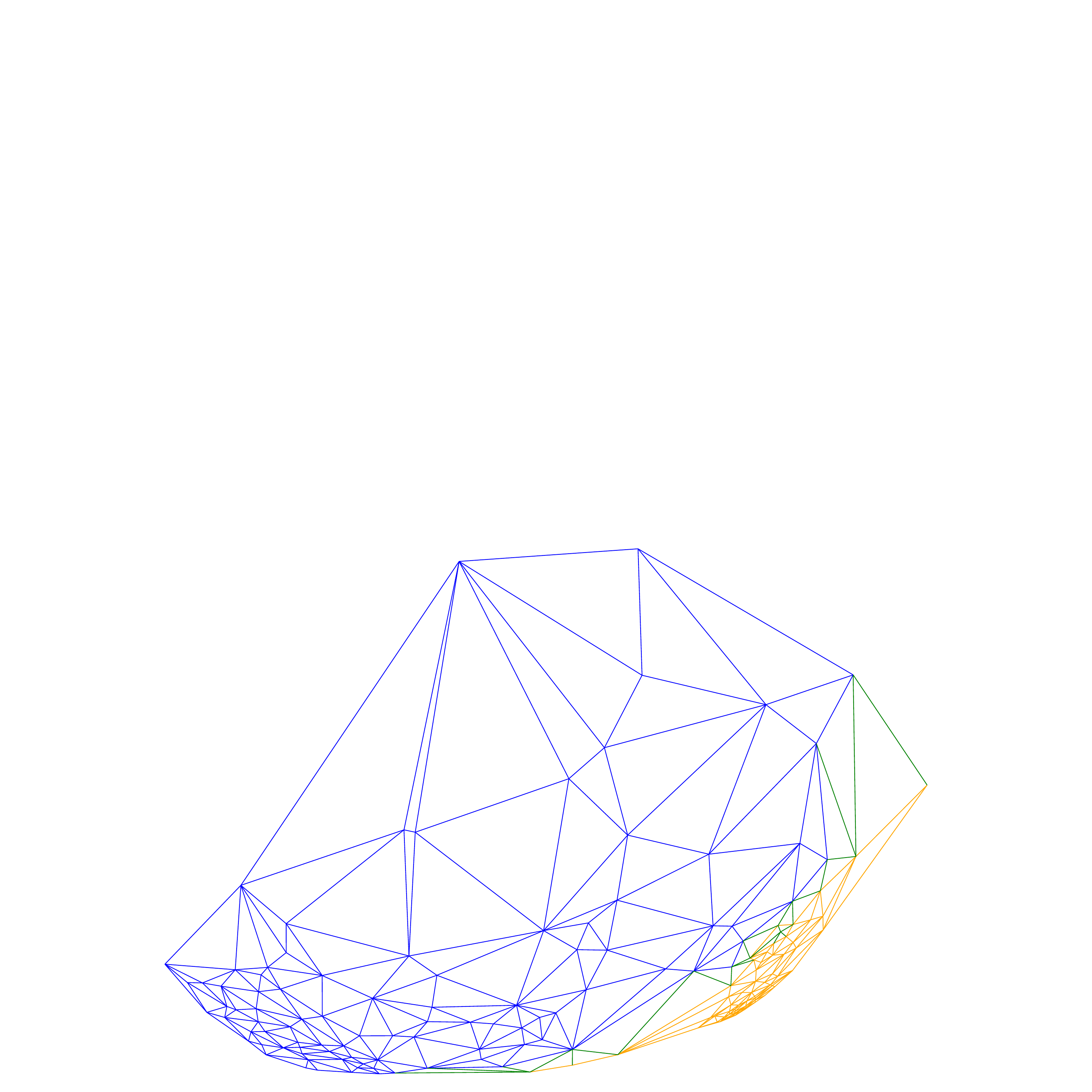}
        \subcaption*{(a)}
    \end{subfigure}
    \begin{subfigure}[b]{.32\linewidth}
        \centering
        \includegraphics[width=\linewidth]{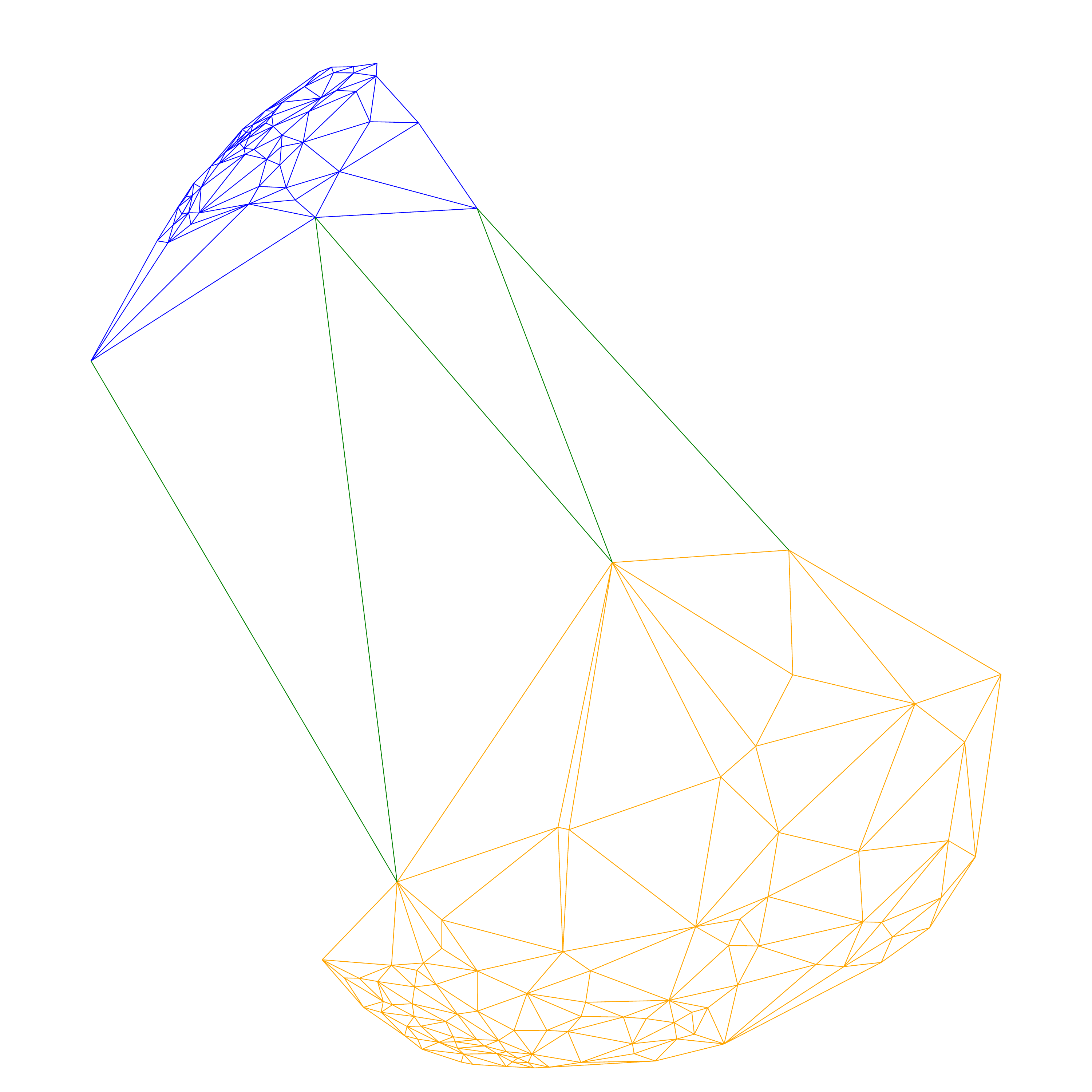}
        \subcaption*{(b)}
    \end{subfigure}
    \begin{subfigure}[b]{.32\linewidth}
        \centering
 \includegraphics[width=\linewidth]{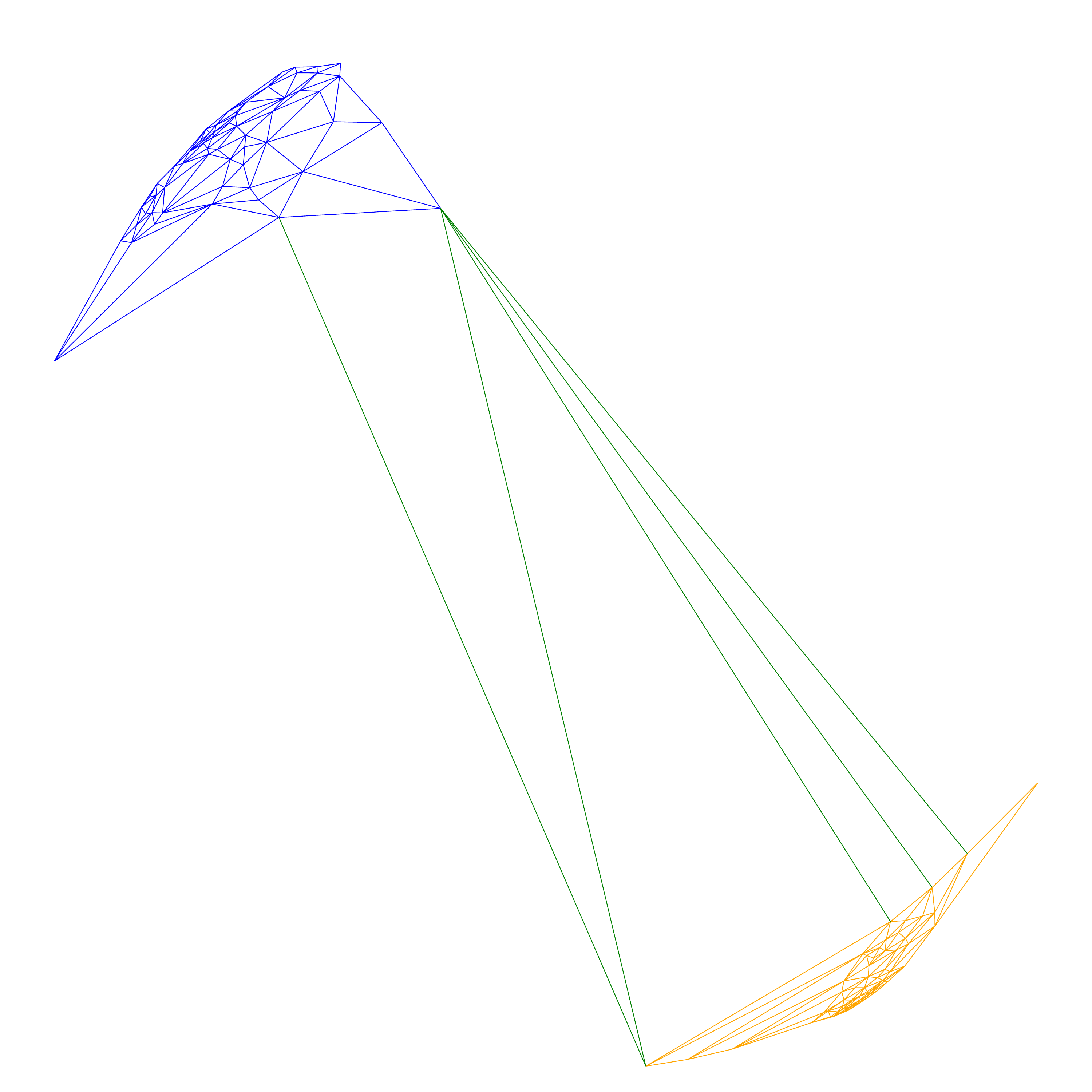}
    \subcaption*{(c)}
    \end{subfigure}

\vspace{2em}
    \begin{subfigure}[b]{.32\linewidth}
        \centering
        \includegraphics[width=\linewidth]{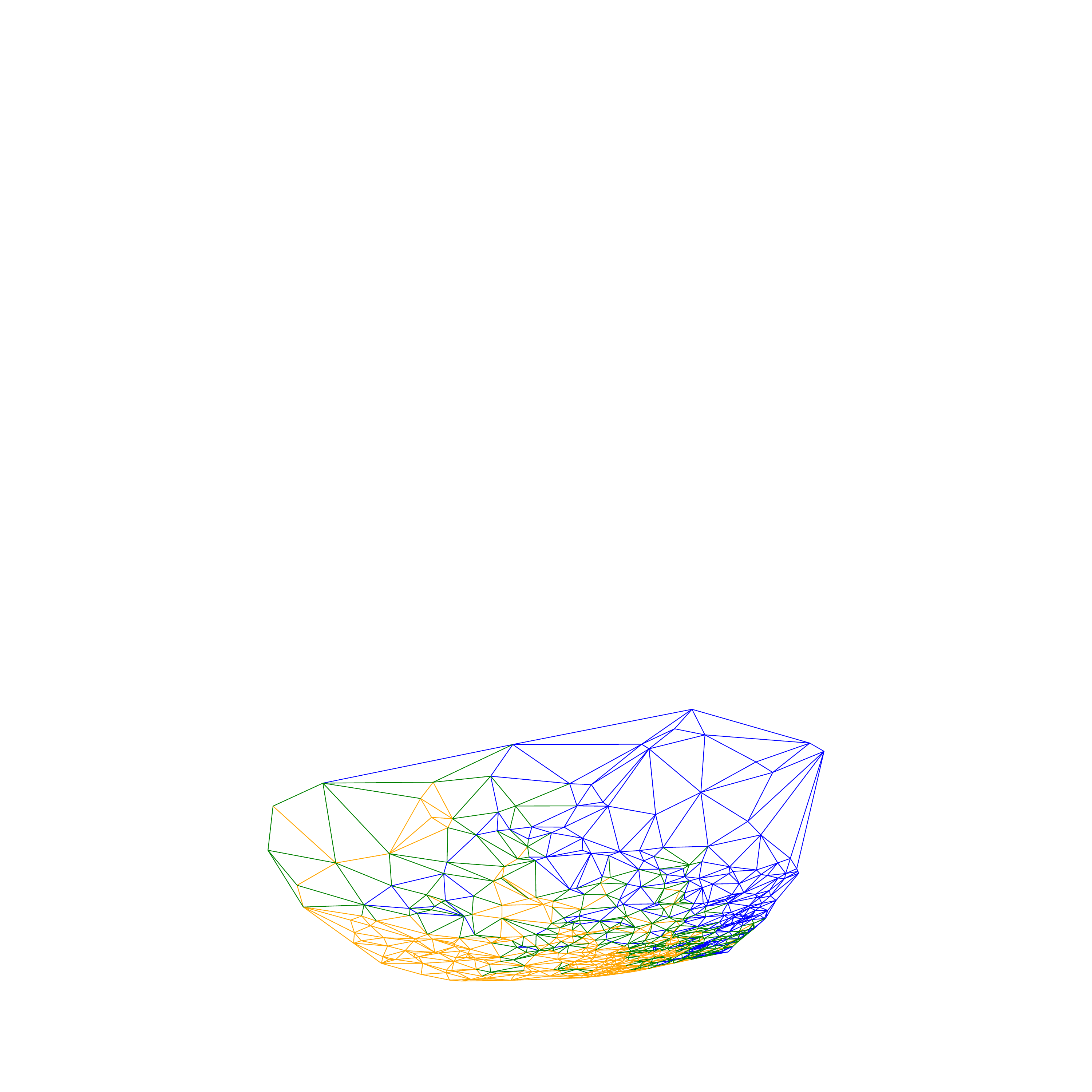}
        \subcaption*{(a)}
    \end{subfigure}
    \begin{subfigure}[b]{.32\linewidth}
        \centering
        \includegraphics[width=\linewidth]{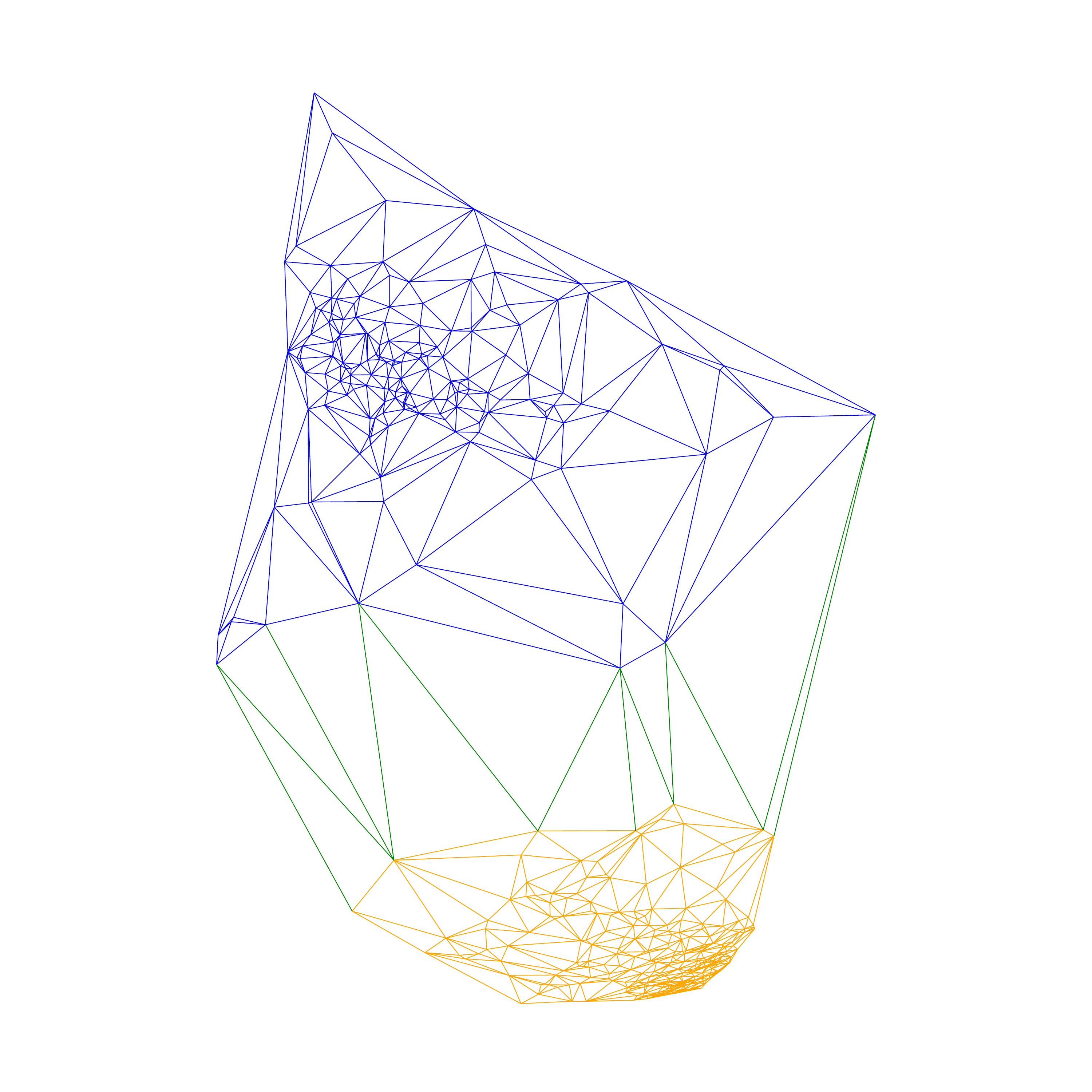}
        \subcaption*{(b)}
    \end{subfigure}
    \begin{subfigure}[b]{.32\linewidth}
        \centering
 \includegraphics[width=\linewidth]{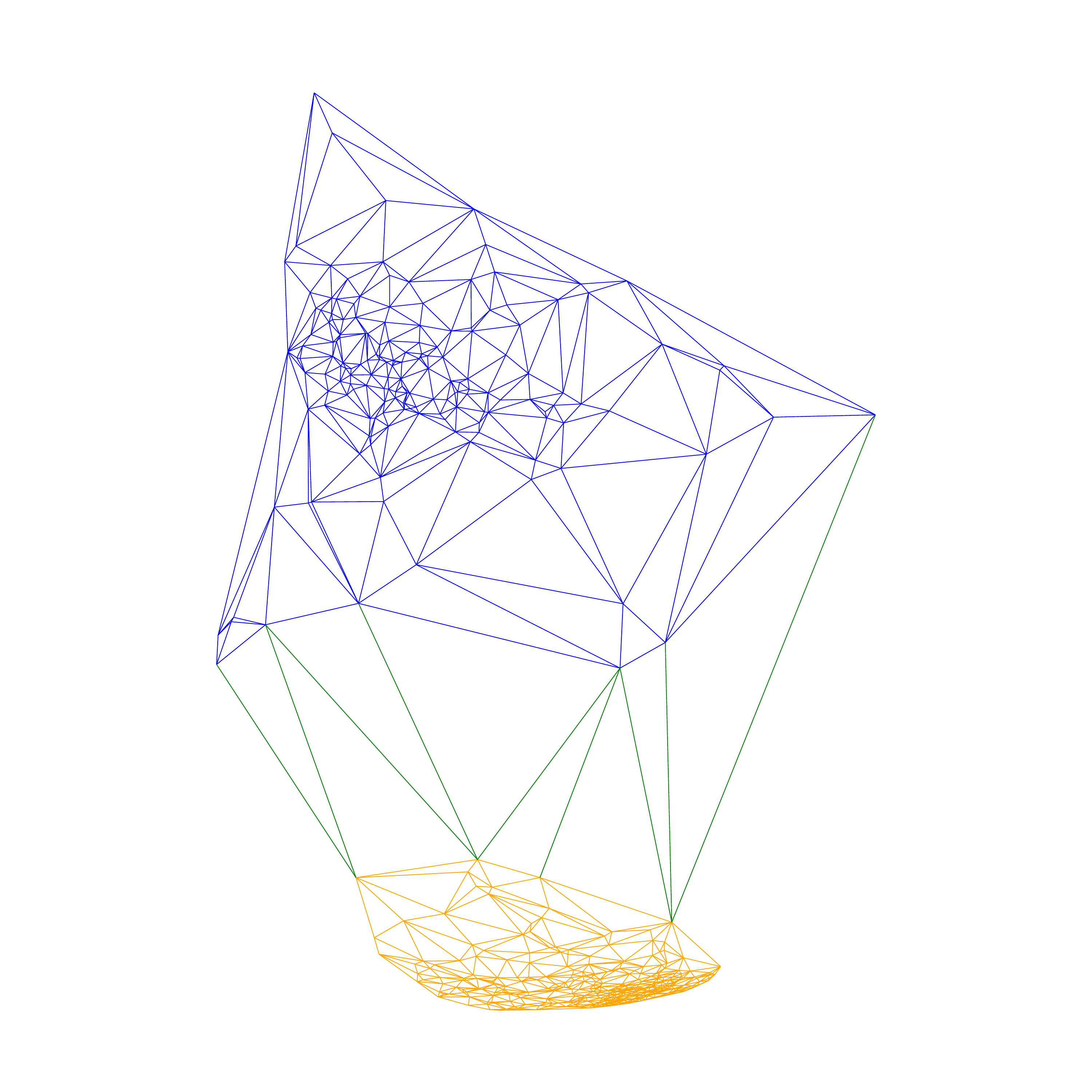}
    \subcaption*{(c)}
    \end{subfigure}

    \vspace{2em}
    \begin{subfigure}[b]{.32\linewidth}
        \centering
        \includegraphics[width=\linewidth]{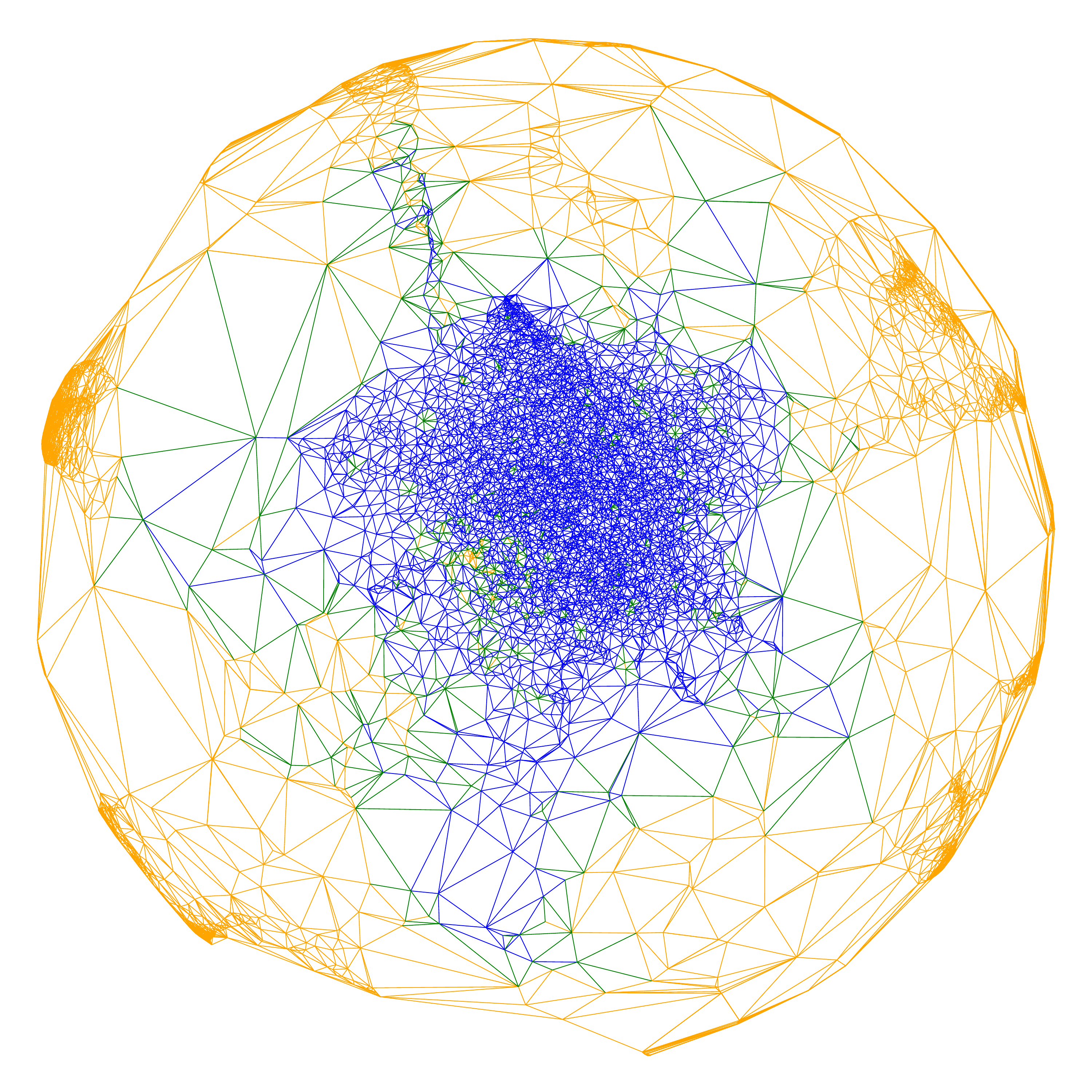}
        \subcaption*{(a)}
    \end{subfigure}
    \begin{subfigure}[b]{.32\linewidth}
        \centering
 \includegraphics[width=\linewidth]{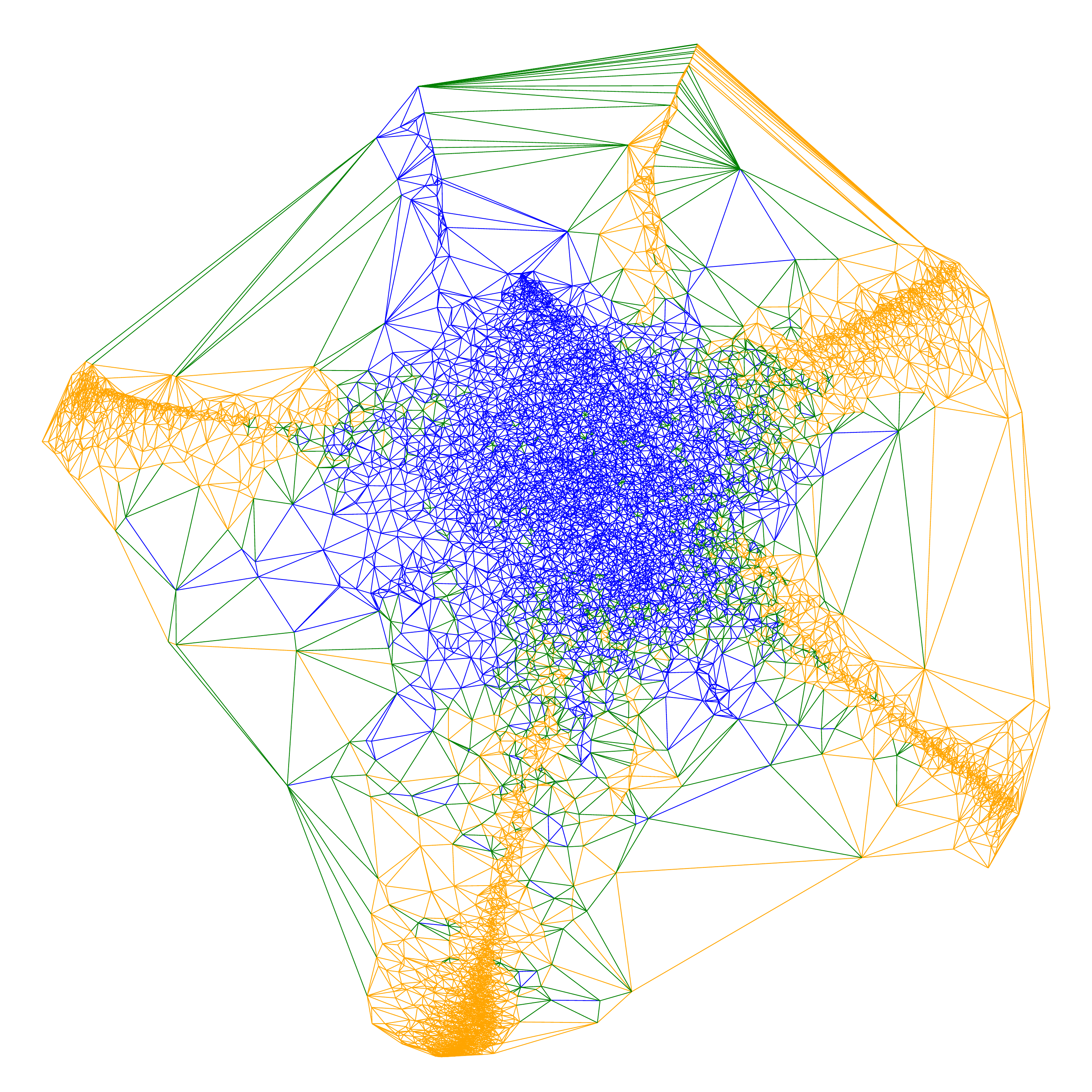}
    \subcaption*{(b)}
    \end{subfigure}
    \begin{subfigure}[b]{.32\linewidth}
        \centering
        \includegraphics[width=\linewidth]{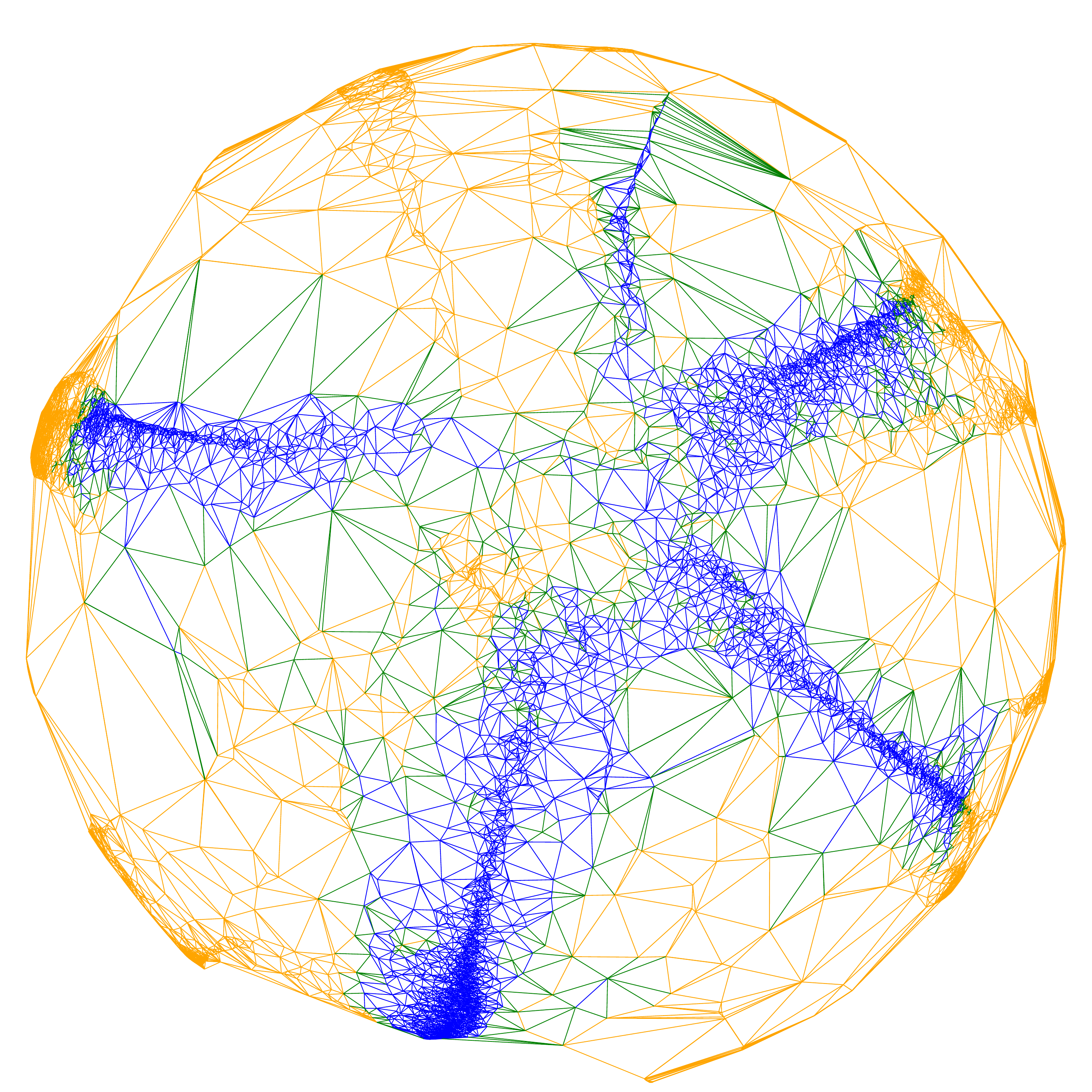}
        \subcaption*{(c)}
    \end{subfigure}
    \caption{HyperDGA visualizations of homogeneous edges (blue and orange) and heterogeneous edges (green) for Experiment 3 in the Klein-Beltrami model. Top: \emph{Olsson}. (a) Mono vs. Gran, (b) HSPC-1 vs. Mono, and (c) HSPC-1 vs. Gran. Center: \emph{Paul}. (a) 14Mo vs. 13Baso, (b) 7MEP vs. 13Baso, and (c) 7MEP vs. 14Mo. Bottom: \emph{Planaria}. (a) neoblasts vs. differentiated, (b) neoblasts vs. progenitors, and (c) progenitors vs. differentiated.}
    \label{fig:experiment3_planaria}
\end{figure*}

\end{document}